\documentclass{article}

\usepackage{graphicx}
\usepackage{booktabs}
\usepackage{multirow}
\usepackage{algorithm}
\usepackage{algorithmic}
\usepackage{amsmath}
\usepackage{tabularx}
\usepackage{textcomp}

\usepackage[dvipsnames,table]{xcolor}
\usepackage{colortbl}
\usepackage{array}
\usepackage{rotating}
\usepackage{natbib}

\usepackage{graphicx}
\usepackage{subcaption}
\usepackage{CJKutf8}
\usepackage{enumitem}

\usepackage{pifont}
\usepackage{adjustbox}

\usepackage{booktabs}
\usepackage{array}
\usepackage{adjustbox}
\usepackage{makecell}

\newcommand{\mypara}[1]{\smallskip\noindent\textbf{#1}}

\usepackage[final]{neurips_2024}
\reportdate{April 2026}
\reportprojectpage{\url{https://starvla.github.io}}

\usepackage[utf8]{inputenc}
\usepackage[T1]{fontenc}
\usepackage{hyperref}
\usepackage{url}
\usepackage{booktabs}
\usepackage{amsfonts}
\usepackage{nicefrac}
\usepackage{microtype}
\usepackage{tocloft}

\makeatletter
\renewcommand{\@makefnmark}{}
\makeatother

\makeatletter
\renewcommand\@makefntext[1]{\noindent #1}
\makeatother

\title{StarVLA-$\boldsymbol{\alpha}$: Reducing Complexity in Vision-Language-Action Systems}

\author{
  Jinhui Ye$^{1, \dagger}$\quad
  Ning Gao$^{2, \dagger}$\quad
  Senqiao Yang$^{3}$\quad
  Jinliang Zheng$^{4}$\quad
  Zixuan Wang$^{1}$\quad
  Yuxin Chen$^{1}$\\
  Pengguang Chen$^{6}$\quad
  Yilun Chen$^{5,\ddagger}$\quad
  Shu Liu$^{6}$\quad
  Jiaya Jia$^{1,6,\ddagger}$\\[0.5em]
  {\small
    $^{1}$HKUST\quad
    $^{2}$XJTU \quad 
    $^{3}$CUHK \quad
    $^{4}$THU \quad 
    $^{5}$Tongyi Lab, Alibaba Group\quad
    $^{6}$SmartMore Ltd. \\
  }
}

\begin{document}

\maketitle
\footnotetext{
    $^{\dagger}$ Equal contribution \quad
    $^{\ddagger}$ Corresponding author
}

\begin{abstract}
Vision-Language-Action (VLA) models have recently emerged as a promising paradigm for building general-purpose robotic agents. However, the VLA landscape remains highly fragmented and complex: as existing approaches vary substantially in architectures, training data, embodiment configurations, and benchmark-specific engineering. In this work, we introduce \textbf{StarVLA-$\boldsymbol{\alpha}$}, a simple yet strong baseline designed to study VLA design choices under controlled conditions. StarVLA-$\alpha$ deliberately minimizes architectural and pipeline complexity to reduce experimental confounders and enable systematic analysis. Specifically, we re-evaluate several key design axes, including action modeling strategies, robot-specific pretraining, and interface engineering. Across unified multi-benchmark training on LIBERO, SimplerEnv, RoboTwin, and RoboCasa, the same simple baseline remains highly competitive, indicating that a strong VLM backbone combined with minimal design is already sufficient to achieve strong performance without relying on additional architectural complexity or engineering tricks. Notably, our single generalist model outperforms $\pi_{0.5}$ by 20\% on the public real-world RoboChallenge benchmark. We expect StarVLA-$\alpha$ to serve as a solid starting point for future research in the VLA regime. Code will be released at \url{https://github.com/starVLA/starVLA}.
\end{abstract}

\section{Introduction}
\label{sec:intro}

\begin{figure}[t]
    \centering
    \includegraphics[width=1\linewidth]{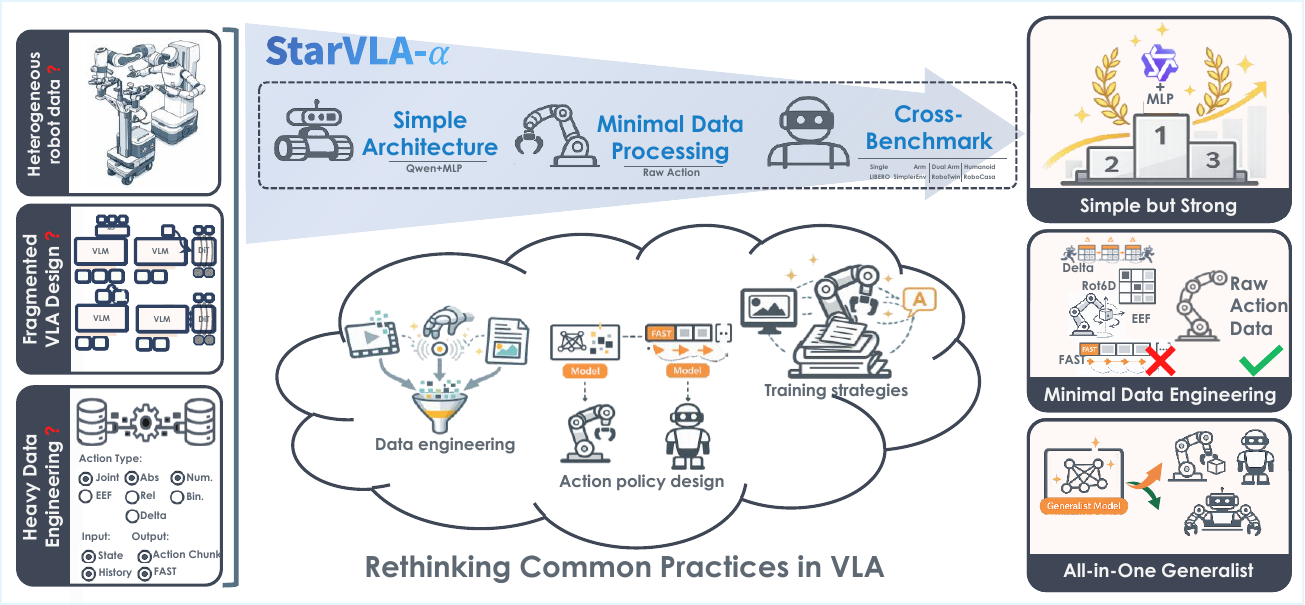}
    \caption{Current VLA systems are difficult to compare due to heterogeneous robot datasets, fragmented architectures, and heavy benchmark-specific engineering. StarVLA-$\alpha$ removes these confounders with a simple VLM-based architecture, minimal data processing, and unified cross-benchmark training. This controlled baseline enables systematic analysis of action modeling, robot pretraining, and interface design, revealing that many commonly adopted complexities provide limited context-dependent benefits.}
    \label{fig:teaser}
\end{figure}

Recent progress in robotic manipulation has been increasingly driven by Vision-Language-Action (VLA) models, which aim to move beyond task-specific policies toward general-purpose robotic agents. Since the introduction of RT-series~\cite{RT-2,RT-1,rth} as robotic foundation models, the field has rapidly evolved by leveraging large foundation models, scaling robot data~\cite{pi_0, robomind, generalistai} and general multimodal supervision~\cite{intelligence2025pi05, RT-2, instructvla, chen2025internvla, ye2026st4vla} to achieve impressive policy transferability and task coverages~\cite{RT-1,RT-2,openvla,octo,black2024pi_0,intelligence2025pi05}. As a result, a growing number of VLA systems demonstrate impressive results across a variety of robotic benchmarks~\cite{simpleenv, Libero, mees2022calvin, Mu_robotwin1, chen2025robotwin2, maniskill2}. Meanwhile, open-source efforts~\cite{openvla, pi_05, bjorck2025gr00t, liu2024rdt, cai2026internvla} have broadened accessibility and accelerated experimentation.

Despite the rapid development of VLA systems, the field still lacks a clear understanding of which components actually drive performance gains. Existing systems vary in model architectures, pre-training data, embodiment configurations, and benchmark-specific fine-tuning, making empirical comparison difficult to interpret. Reported improvements are often entangled with dataset choices, preprocessing pipelines, and benchmark-specific engineering, obscuring whether gains arise from modeling innovations or experimental variation. In contrast to vision-language modeling (VLM), where training practices have gradually converged toward standardized recipes~\cite{llavaov, instructblip, llava_next}, VLA research remains highly fragmented. Establishing clearer methodological consensus is therefore increasingly important for guiding future progress in the field.

However, reaching methodological consensus is a challenging and long-standing problem due to substantial heterogeneity across the VLA pipeline as shown in Fig.~\ref{fig:teaser}. First, pre-training data and embodiment configurations vary substantially across studies. Rapid evolution of robotic platforms and teleoperation pipelines has led to heterogeneous datasets with incompatible interfaces, action spaces, and normalization schemes~\cite{openvla, liu2024rdt}. Robot embodiments span single-arm manipulators such as Franka and UR5~\cite{franka, ur5}, wheeled dual-arm systems~\cite{galbot, galaxea, agibot}, and humanoid robots~\cite{fourier_gr1, unitree, agibot}, accompanied by differences in camera viewpoints and end-effectors, further entangling modeling choices with embodiment-specific preprocessing. 
Second, modeling and training strategies lack consensus. Existing VLA systems adopt diverse combinations of vision towers, language backbones, and action experts~\cite{octo, openvla, cogact, roboflamingo, pi_0, pi_05, walloss2025}, while design choices such as action parameterization and normalization for continuous robot states and controls remain poorly understood. Third, varied evaluation practices complicate comparison. Benchmark-specific hyperparameter tuning, dataset splits, and action chunking strategies are often required to achieve strong performance~\cite{simpleenv, Libero, Mu_robotwin1, chen2025robotwin2, robocasa, behavior1k}, and strong in-benchmark results do not necessarily translate to robustness under broader distribution shifts~\cite{pumacay2024colosseum, robocasa, gao2025genmanip}. 

To demystify the essential components of VLA systems, we propose \textbf{StarVLA-$\boldsymbol{\alpha}$} upon the infrastructure of StarVLA~\cite{starvla}, a simple yet strong baseline that serves as a starting point for systematically studying existing VLA paradigms. It is explicitly designed to reduce experimental confounding and isolate modeling effects. Rather than introducing additional architectural complexity, we deliberately minimize structural variations by employing a pre-trained VLM backbone (Qwen3-VL) without robot-specific pre-training or sophisticated action engineering. We follow official evaluation protocols and avoid benchmark-specific tuning to ensure controlled and reproducible comparisons. The objective is not architectural novelty but methodological clarity: by controlling major sources of variation, StarVLA-$\alpha$ provides a \textit{controlled substrate} for reassessing widely adopted VLA design choices under comparable conditions.

Under this controlled setting, a strong VLM-based baseline matches or exceeds recent VLA systems while keeping the backbone, training data, and training settings identical. Under controlled conditions, we examine the necessity of common VLA design choices along three axes: action head design, robot-specific pretraining, and data/interface engineering. Keeping the backbone, data scale, and training protocol identical, we compare several canonical VLM-to-VLA instantiations within a unified pipeline, including discrete token-based autoregressive decoding (FAST-style), direct continuous action regression with a lightweight MLP head (OpenVLA-OFT-style), diffusion/flow-matching based continuous action generation ($\pi_0$-style), and dual-system designs that couple a VLM with a separate low-level action module (GR00T-style), finding that simple MLP action header remains highly competitive while more complex designs provide only scenario-dependent gains (see Sec.~\ref{sec:abl_action_header}). Robot pretraining by incorporating large-scale action data~\cite{open_x_embodiment, interndata-a1} is re-assessed. We observe that heterogeneous pretraining may impair cross-embodiment generalization and that domain-aligned data yields conditional rather than overall improvements (Sec.~\ref{sec:abl_pretraining}). Finally, we revisit common engineering choices (e.g. auxiliary inputs, action output modeling). Overall, {removing major confounders reveals that architectural and engineering complexity offers limited and context-dependent gains} (Sec.~\ref{sec:abl_data_tricy}).

To mitigate potential benchmark-specific bias in single-benchmark evaluation, we further assess robustness under broader generalization regimes. We jointly train a unified model across LIBERO~\cite{Libero}, SimplerEnv~\cite{ovdet}, RoboTwin 2.0~\cite{chen2025robotwin2}, and RoboCasa-GR1~\cite{robocasa, bjorck2025gr00t} without benchmark-specific adaptation, using unified action padding across embodiments (Sec.~\ref{sec:all_in_one}). Under this multi-benchmark setting, the same simple baseline remains competitive, and in several cases superior to task-specific models. These results indicate that strong backbone initialization and unified training can support cross-task and cross-embodiment generalization without requiring additional architectural complexity.

Our contributions are summarized as follows:
\begin{itemize}[leftmargin=0.1in]
    \item We present a simple yet strong VLA baseline that removes key confounders, showing that a streamlined VLM design can reach leading performance on four benchmarks spanning five embodiments.
    \item Under controlled backbone, data, and training settings, we systematically re-evaluate common VLA design choices and find that added architectural/data engineering complexity yields smaller and more context-dependent gains than often assumed.
    \item We further demonstrate that a single generalist model trained jointly across benchmarks, without task-specific adaptation, can generalize across tasks and embodiments, supported by strong initialization and a standardized pipeline. 
    % Notably, RoboChallenge~\yilun{TODO}
\end{itemize}

\section{StarVLA-$\boldsymbol{\alpha}$}
\label{sec:clean-vla}

Since the introduction of RT-1~\cite{RT-1} in 2022, Vision–Language–Action (VLA) research has pursued general-purpose embodied agents built on foundation models. Along the way, the community has explored many design dimensions—vision backbones (e.g., SigLIP~\cite{siglip}, DINO~\cite{oquab2023dinov2}, CLIP~\cite{clip}), action heads (discrete tokens, continuous regression, diffusion, flow matching), and action/data pipelines (delta vs.\ relative actions; embodiment-specific preprocessing across eef/joint/6D pose). While these choices have driven steady gains, they have also fragmented the field: systems often become complex, hard to reproduce, and tightly tuned to benchmark-specific details, which can hurt transfer to new embodiments.

\begin{figure}[t]
    \centering
    \includegraphics[width=1\linewidth]{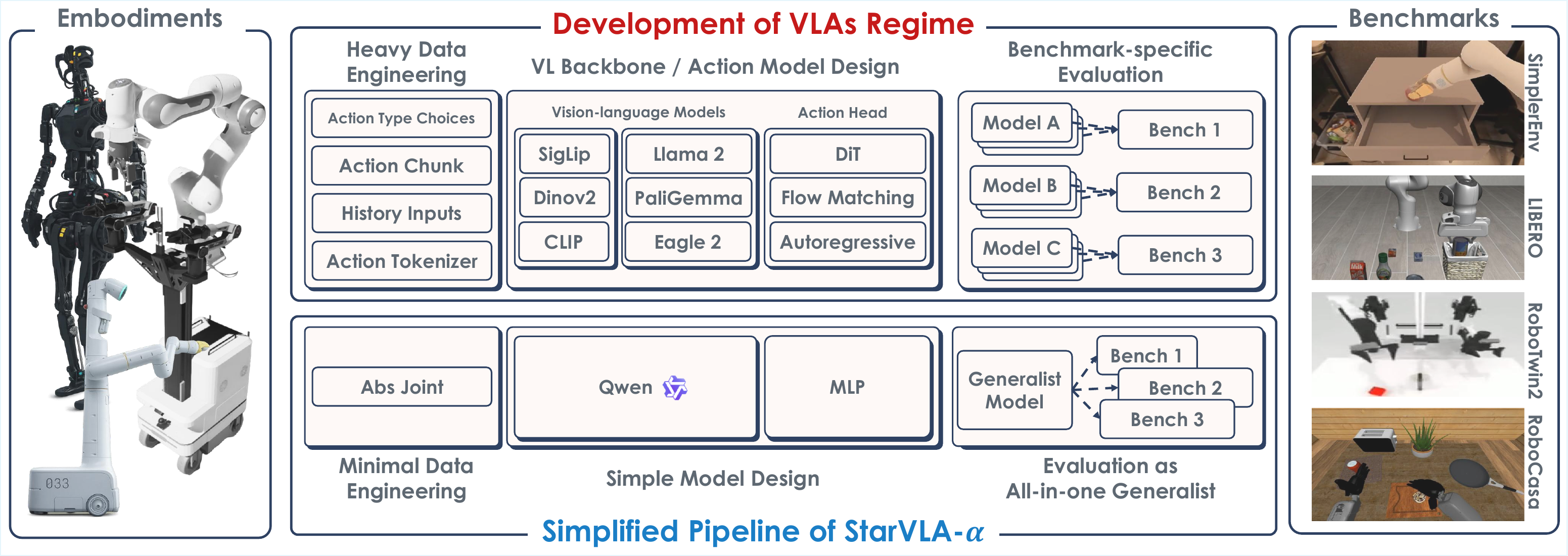}
    \caption{\textbf{Overview of StarVLA-$\boldsymbol{\alpha}$}. We use a unified VLM backbone (Qwen3-VL) with minimal preprocessing and a lightweight MLP action head. This simple setup avoids specialized vision encoders, benchmark-specific data pipelines, and complex action heads, while enabling consistent training and evaluation across diverse benchmarks.}
    \label{fig:method}
\end{figure}

Against this backdrop, we ask a simple question: \textbf{can we cut through this complexity?} Specifically, we test whether a strong VLM backbone can deliver competitive performance without elaborate architectures or heavy data engineering. To study this, we build a clean, transparent, and robust VLA baseline from scratch (Fig.~\ref{fig:method}).

\subsection{A Simple and Unified VLA Framework}
\label{sec:framework}

Our framework is guided by a minimal-sufficiency hypothesis: a strong VLM paired with a lightweight action head captures most of the benefits commonly attributed to more complex designs.
Here, ``clean'' refers to two aspects: minimal data processing and a simple architecture.

\mypara{Minimal data processing.}
To promote generalization across diverse robot embodiments and benchmarks, we use a single minimal data pipeline shared across all environments. The model takes raw RGB images and the provided language instructions as input, without benchmark-specific engineering or custom formatting. We normalize actions using the training split only (zero mean, unit variance). For evaluation, we follow each benchmark’s official protocol. This unified preprocessing makes the framework directly applicable to new robot embodiments and benchmarks without additional adaptation.

\mypara{Clean architecture.}
We follow common practice and couple a VLM backbone with a lightweight action head for continuous action prediction.  We instantiate the backbone with the {Qwen} family of models~\cite{qwen2vl, qwen3vl}, specifically Qwen3-VL. We choose Qwen for two reasons: (i) it is a widely adopted open-source VLM with strong community support; and (ii) its unified design natively processes both vision and language inputs, avoiding the need to separately select and combine vision encoders (e.g., CLIP~\cite{clip}, SigLIP~\cite{siglip}). On top of the VLM, we attach a simple MLP action head that reads the hidden state of a designated action token and regresses a chunk of continuous actions. The modular design also allows us to swap in alternative VLM backbones or action heads with minimal changes.

\mypara{Unified benchmark integration.}
To enable systematic evaluation, we integrate a diverse suite of manipulation benchmarks (e.g., LIBERO, SimplerEnv, RoboTwin 2.0, and RoboCasa-GR1) into a unified pipeline without benchmark-specific design. For each benchmark, we strictly follow its original data and evaluation protocols, applying only our minimal processing while keeping the action representation consistent. We confine heterogeneity to thin adapters that standardize observation formats, action interfaces, and evaluation entry points. As a result, the same model and training recipe run across all benchmarks without customization, and the framework remains easy to extend to new benchmarks. This setup also enables a more general evaluation regime: \textbf{training a single model jointly across all benchmarks}. We refer readers to Sec.~\ref{sec:all_in_one} for detailed results in this unified multi-benchmark setting.

\subsection{Experimental Setup}

We evaluate our models on a diverse set of widely used manipulation benchmarks: \textbf{LIBERO}~\cite{Libero}, \textbf{SimplerEnv}~\cite{simpleenv}, the dual-arm benchmark \textbf{RoboTwin 2.0}~\cite{chen2025robotwin2}, and the humanoid benchmark \textbf{RoboCasa-GR1}~\cite{robocasa, bjorck2025gr00t}. Benchmark details are provided in Appendix~\ref{sec:benchmark_detail}.

\mypara{Baselines.} We compare StarVLA-$\alpha$ against several representative VLA methods: FAST~\cite{pertsch2025fast}, OpenVLA-OFT~\cite{openvla}, $\pi_0$~\cite{pi_0}, and GR00T-N1.6~\cite{bjorck2025gr00t}. These prior methods are typically trained separately on each benchmark with their own task-specific data processing. For our approach, we consider two training protocols: (1) \textit{Specialist training}, where we train StarVLA-$\alpha$ independently on each benchmark's training set using our unified minimal data pipeline 
, and (2) \textit{Generalist training}, where we merge all benchmarks' data into a single training set and train a single model. 
We note that the Generalist model represents a large-scale unified training scenario and is included for completeness, \textit{for direct comparisons under similar computational budgets, we focus primarily on Specialist training}.
In the unified setting, actions from different robots are simply padded to a maximum dimension (here 32) with zeros, requiring no per-task engineering; more details are provided in Sec.~\ref{sec:all_in_one}. Training and implementation details are given in Appendix~\ref{sec:training_detail}.

\begin{table}[tbp]
\centering
\caption{\textbf{Performance comparison of StarVLA-$\alpha$ with existing VLAs. }
$^*$ indicates that both clean and random data are used for training. Default StarVLA-$\alpha$ represents multiple models trained separately on each benchmark-specific dataset, while Generalist represents a single model jointly trained across all datasets.}
\label{tab:main_results}
\small
\begin{adjustbox}{width=\linewidth}
\begin{tabular}{lccccc ccc ccc c}
\toprule
\multirow{2}{*}{Method}
& \multicolumn{5}{c}{LIBERO}
& \multicolumn{3}{c}{SimplerEnv}
& \multicolumn{3}{c}{RoboTwin 2.0}
& \multirow{1}{*}{RoboCasa-GR1} \\
\cmidrule(lr){2-6} \cmidrule(lr){7-9} \cmidrule(lr){10-12} \cmidrule(lr){13-13}
& Spatial & Object & Goal & Long & avg
& WidowX & Google VA & Google VM
& clean & clean$^*$ & random$^*$
& (avg of 24 tasks) \\
\midrule
\textbf{Specialist}\\
% 过去的方法 with 各式各样的 技巧\\
OpenVLA-OFT   & 97.6 & 98.4 & 97.9 & 94.5 & 97.1 & 31.3 & 54.3 & 63.0   & --   & --   & --   & -- \\
$\pi_0$        & 96.8 & 98.8 & 95.8 & 85.2 & 94.1 & 27.1 &  54.8  & 58.8   & 46.42 & 65.9 & 58.4 & -- \\
$\pi_0$+FAST & 96.4 & 96.8 & 88.6 & 60.2 & 85.5 &  39.5 & 60.5 & 61.9 & --   & --   & --   & -- \\
$\pi_{0.5}$        & 98.8 & 98.2 & 98.0 & 92.4 & 96.9 & 46.9  & 68.4 & 72.7 &  \textbf{60.2} & 82.7 & 76.8 & 37.0 \\
GR00T-N1.6    & 97.5 & 98.5 & 97.5 & \textbf{94.4} & 97.0 & 62.0 & 65.3 & 67.7 &  --  & --   & -- & 47.6 \\
StarVLA-$\alpha$    & \textbf{99.0} & \textbf{99.8} & 98.5 & 94.1 & \textbf{98.8} & 64.6 & \textbf{70.2} & \textbf{76.0} & 50.3 & 88.2 & \textbf{88.3} & 53.8 \\
\midrule
StarVLA-$\alpha$ (Generalist) & 98.7 & 99.7 & \textbf{98.6} & 94.2 & 97.8 & \textbf{65.2} & 69.8 & 74.3 & -- & \textbf{88.7} & 87.8 & \textbf{57.3}  \\

\bottomrule
\end{tabular}
\end{adjustbox}
\end{table}

\subsection{Main Results}

As shown in Table~\ref{tab:main_results}, StarVLA-$\alpha$ performs strongly across all benchmarks relative to prior VLA methods. On LIBERO, StarVLA-$\alpha$ achieves an average success rate of 98.8\%, outperforming all previous approaches. On SimplerEnv, it exceeds the best existing method by a substantial margin (e.g., +6.8\% on Google VM). On the more challenging dual-arm and humanoid settings, StarVLA-$\alpha$ reaches up to 53.8\% success, highlighting the strength of a capable VLM backbone even with a lightweight action head.

Moreover, under a unified generalist training setup, we find that training a single model on diverse data achieves competitive per-benchmark performance while notably improving on challenging benchmarks such as RoboCasa-GR1 (Sec.~\ref{sec:all_in_one}).

Together, these results support our central hypothesis: \textit{a strong VLM, paired with a straightforward action head and minimal data preprocessing, can deliver highly competitive performance.} More broadly, they suggest a practical way to reduce the field’s growing complexity: fix the backbone, standardize the data pipeline, and avoid task-specific engineering. This approach yields a strong, reproducible baseline that can serve as a solid foundation for future work.

\section{Rethinking Common Practices in VLA Systems}
\label{sec:rethink}

As described in Sec.~\ref{sec:clean-vla}, StarVLA-$\alpha$ baseline is intentionally simple: it pairs a strong VLM (Qwen3-VL) with a lightweight MLP action head, uses minimally processed data, and introduces no state inputs, history frames, or additional pretraining. Despite this minimal design, StarVLA-$\alpha$ achieves state-of-the-art results across multiple benchmarks and substantially outperforms prior methods. This finding motivates a natural question: \textbf{when a strong backbone is available, what actually drives VLA performance?} In this section, we systematically analyze three commonly emphasized design choices: action head architecture, action-specific pretraining, and data engineering.

\subsection{Do Different Action Head Designs Matter?}
\label{sec:abl_action_header}

\begin{figure}[t]
    \centering
    \includegraphics[width=1\linewidth]{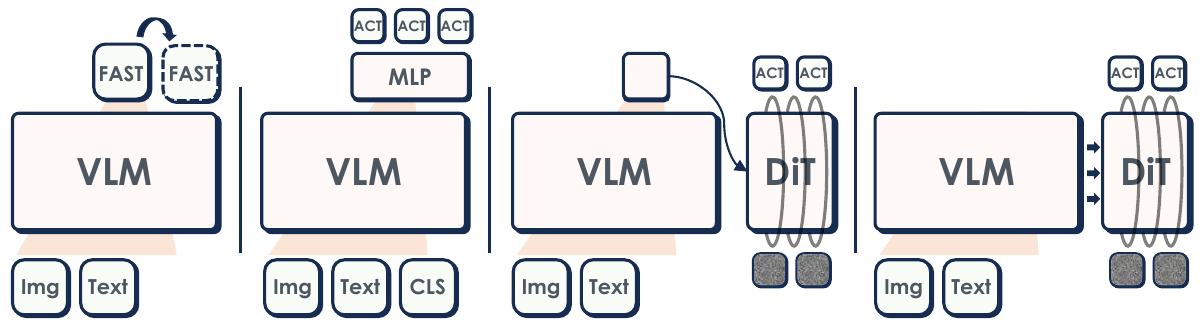}
    \caption{\textbf{Action expert designs on StarVLA-$\alpha$.} From left to right: StarVLA-$\alpha$-FAST, StarVLA-$\alpha$ (MLP regression), StarVLA-$\alpha$ -GR00T (dual-system flow matching), and StarVLA-$\alpha$-PI (diffusion-style flow matching).}
    \label{fig:model}
\end{figure}

\mypara{Motivation.} Given that our simple MLP head already delivers strong performance, we ask \textit{whether more complex action heads (e.g. fast token predictors, diffusion models, or dual-system designs) provide additional benefits} when paired with the same strong VLM backbone. Prior comparisons have often been confounded by differences in backbones and training recipes; our unified framework enables us to isolate the effect of the action head itself.

\mypara{Implementation details.} As shown in Fig.~\ref{fig:model}, we evaluate four commonly used designs:
(1) \textbf{StarVLA-$\boldsymbol{\alpha}$-FAST}: Discrete action prediction via an autoregressive FAST tokenizer, similar to $\pi_0$-FAST~\cite{pertsch2025fast}.
(2) \textbf{StarVLA-$\boldsymbol{\alpha}$}: Continuous action regression with a lightweight MLP head applied to dedicated action tokens, following OpenVLA-OFT~\cite{openvla-oft}.
(3) \textbf{StarVLA-$\boldsymbol{\alpha}$-GR00T}: A dual-system architecture where the VLM serves as System~2 (high-level reasoning) and a flow-matching module acts as System~1 for action execution, following GR00T N1.5~\cite{bjorck2025gr00t}.
(4) \textbf{StarVLA-$\boldsymbol{\alpha}$-$\pi$}: Diffusion-style continuous action prediction using a flow-matching expert, analogous to $\pi_0$~\cite{pi_0}.

\begin{table}[tbp]
\centering
\caption{\textbf{Performance comparison across action head designs.}} 
\label{tab:arch_results}
\small
\begin{adjustbox}{width=\linewidth}
\begin{tabular}{lccccc ccc ccc c}
\toprule
\multirow{2}{*}{Method}
& \multicolumn{5}{c}{LIBERO}
& \multicolumn{3}{c}{SimplerEnv}
& \multicolumn{3}{c}{RoboTwin 2.0}
& \multirow{1}{*}{RoboCasa-GR1} \\
\cmidrule(lr){2-6} \cmidrule(lr){7-9} \cmidrule(lr){10-12} \cmidrule(lr){13-13} 
& Spatial & Object & Goal & Long & avg
& WidowX & Google VA & Google VM
& clean & clean$^*$ & random$^*$
& (avg of 24 tasks)\\
\midrule

StarVLA-$\alpha$   & \textbf{99.0} & \textbf{99.8} & \textbf{98.5} & 94.1 & \textbf{98.8} & 64.6 & 70.2 & 76.0 & {50.3} & \textbf{88.2} & 88.3 & \textbf{53.8} \\
StarVLA-$\alpha$-FAST  & 98.3 & 98.4 & 97.3 & 91.6 & 97.8  & 35.6 & 58.8   & 60.1   & 46.4   & 72.5   & 83.2   & 45.0 \\
StarVLA-$\alpha$-GR00T & 98.9 & 99.6 & 98.4 & \textbf{95.3} & 98.7 & 65.3 & 70.7   & 75.3   & 48.8   & 88.0   & 88.5   & 52.8 \\
StarVLA-$\alpha$-$\pi$    & 98.0 & 99.2 & 98.2 & 93.6 & 98.1  & \textbf{65.9} & \textbf{72.8}   & \textbf{76.6}   & \textbf{50.8}   & 88.1   & \textbf{88.8}   & 48.9 \\

\bottomrule
\end{tabular}
\end{adjustbox}
\end{table}

\begin{table}[t]
\centering
\caption{\textbf{Effects of additional robotic data pretraining.}}
\label{tab:pretrain_ablation}
\begin{adjustbox}{max width=\linewidth}
\begin{tabular}{lccccccc}
\toprule
\multirow{2}{*}{\textbf{Mid-Pretraining}} & \multirow{2}{*}{\textbf{Traj. Num.}} & \multicolumn{3}{c}{\textbf{RoboTwin-Clean}} & \multicolumn{3}{c}{\textbf{RoboCasa-GR1}} \\
\cmidrule(lr){3-5} \cmidrule(lr){6-8}
& & Clean $50\times50$ & +Random $\times100$ & +Random $\times500$ & $24\times10$ & $24\times100$ & $24\times1000$ \\
\midrule
% Random init & 20.8 & 30.6 & 77.8 & 0.0 & 5.6 & 17.8 \\
StarVLA-$\alpha$ & - & 50.3 & 78.5 & 88.2 & \textbf{9.8} & \textbf{39.4} & \textbf{53.8} \\
+ OXE & 232.6k & 30.2 & 40.6 & 83.6 & 1.2 & 17.7 & 27.8 \\
+ InternData-A1 & 630k & 63.6 & 80.4 & 88.6 & 2.8 & 27.6 & 35.4 \\
+ RoboTwin-Rand & 25k & \textbf{79.7} & \textbf{84.1} & \textbf{88.8} & 2.2 & 27.3 & 33.3 \\
\bottomrule
\end{tabular}
\end{adjustbox}
\end{table}
\mypara{Main results. }
As shown in Table~\ref{tab:arch_results}, we compare four action head designs across several settings. Continuous action prediction consistently outperforms discrete action prediction (StarVLA-$\alpha$-FAST) on nearly all benchmarks. Among the continuous-action variants, however, the three action heads achieve comparable performance. In particular, all methods reach over 98\% success on LIBERO and around 65\% on WidowX. Notably, the simplest design, StarVLA-$\alpha$, achieves 53.8\% on RoboCasa-GR1.

\mypara{Takeaway. }
These results suggest two key observations: (1) continuous action prediction is critical for strong performance and consistently outperforms discrete token-based approaches; and (2) given a powerful VLM, \textbf{the choice of continuous action head has limited impact}. Consequently, a lightweight MLP head serves as a simple, efficient, and competitive default. This result indicates that additional architectural complexity in the action head is unnecessary when the underlying VLM backbone is sufficiently strong.

\subsection{Does existing action-specific pretraining matter?}
\label{sec:abl_pretraining}
\mypara{Motivation.} Most existing VLA models perform large-scale action-specific pretraining before fine-tuning on downstream tasks. For instance, OpenVLA uses the Open X-Embodiment (OXE) dataset~\cite{openvla}, $\pi_{0.5}$ leverages diverse robot and multimodal data~\cite{pi_05}, and GR00T relies on large-scale simulation datasets~\cite{bjorck2025gr00t}. Such pretraining is widely regarded as important for strong performance. However, our StarVLA-$\alpha$ baseline, built solely on a pretrained VLM (Qwen3-VL-4B) and without any action-specific data, already achieves competitive results. This observation raises a key question: \textit{given a strong VLM backbone, does additional action-specific pretraining provide further benefits?}

\mypara{Experimental setups.}
To answer this question, we use the StarVLA-$\alpha$ architecture and keep all hyperparameters fixed. We compare four pretraining settings and evaluate them on RoboCasa-GR1 and RoboTwin:
(1) \textbf{StarVLA-$\boldsymbol{\alpha}$ (VLM-based):} No additional pretraining; the pretrained Qwen3-VL model is directly fine-tuned on task-specific data.
(2) \textbf{+OXE:} The pretrained Qwen3-VL model is first trained on the OXE dataset~\cite{open_x_embodiment} and then fine-tuned on the task-specific data.
(3) \textbf{+InternData-A1:} The pretrained Qwen3-VL model is first trained on the InternData-A1 dataset~\cite{interndata-a1},  which shares overlapping embodiments (and partially aligned action interfaces) with RoboTwin, and then fine-tuned on the task-specific data.
(4) \textbf{+RoboTwin-Rand:} Qwen3-VL model is pre-trained on RoboTwin randomized data~\cite{chen2025robotwin2} (within the same domain) and then fine-tuned on the task-specific data.

\mypara{Results.}
As shown in Table~\ref{tab:pretrain_ablation}, the StarVLA-$\alpha$ baseline achieves near-best performance when sufficient task-specific data is available, reaching 88.2 and 53.8 on RoboTwin 2.0 and RoboCasa-GR1, respectively. Adding large-scale pretraining data does not consistently improve performance; out-of-domain data such as OXE can even degrade results. Although pretraining with InternData-A1 or RoboTwin data improves RoboTwin performance, particularly in low-data regimes, it still reduces performance on RoboCasa, suggesting that even in-domain gains may not transfer across embodiments or tasks.

\mypara{Takeaway.} \textbf{Additional action-specific pretraining can improve performance when the pretraining data closely matches the target task, but it may hurt generalization to unseen domains.} A strong VLM baseline already provides a solid foundation; further pretraining can therefore act as a double-edged sword and should be applied with caution.

\subsection{Is Data Engineering Necessary?}
\label{sec:abl_data_tricy}

\mypara{Motivation. }
Beyond architecture and pretraining, many VLA models rely on various data engineering techniques to improve performance. These include perception-related inputs, such as proprioceptive states and stacked history frames, as well as action representations, such as absolute, delta, or relative actions. While widely used, the necessity of these techniques remains unclear, particularly when a strong VLM backbone is available. In this section, we systematically examine a set of common data engineering choices within StarVLA-$\alpha$ framework. We evaluate them across multiple benchmarks and data scales to determine whether they yield consistent improvements.

\mypara{Experimental setup.}
We study four commonly used data engineering choices:
(1) \textbf{Proprioception~\cite{pi_0, bjorck2025gr00t}:} adding robot joint states as input, concatenated with VLM features before the action head.
(2) \textbf{History frames~\cite{li2025cronusvla}:} stacking the previous two frames to provide temporal context.
(3) \textbf{Delta action~\cite{feng2026demystifying}:} predicting relative changes from the current joint position.
(4) \textbf{Relative action~\cite{feng2026demystifying}:} predicting actions in a reference coordinate frame (e.g., end-effector–centric).

Each modification is applied to StarVLA-$\alpha$ while keeping all other training hyperparameters unchanged. We report results on three representative benchmarks: LIBERO (average over four tasks), RoboTwin 2.0, and RoboCasa-GR1 under different data scales. For RoboTwin 2.0, the data regimes include Clean $50\times50$, +Random $100$, and +Random $500$. For RoboCasa, we evaluate with $24\times10$, $24\times100$, and $24\times1000$ demonstrations.

\mypara{Results.}
As shown in Table~\ref{tab:data_engineering}, when the dataset is small, for example, Clean $50\times50$ on RoboTwin 2.0 or $24\times10$ on RoboCasa-GR1, certain data engineering techniques provide modest improvements. However, once sufficient task-specific data is available, these techniques offer little additional benefit and perform similarly to the baseline without data engineering.

\mypara{Takeaway.}
When built upon a strong VLM and a clean codebase, \textbf{data engineering techniques can offer modest benefits when task-specific data is limited}. However, their impact \textbf{becomes negligible once sufficient task-specific data is available}.

\begin{table}[t]
\centering
\caption{\textbf{Ablation study on data engineering across benchmarks and data scales.}}
\label{tab:data_engineering}
\begin{adjustbox}{max width=\linewidth}
\begin{tabular}{lccccccc}
\toprule
\multirow{2}{*}{\textbf{Mid-Pretraining}} 
& \multirow{2}{*}{\textbf{LIBERO}}  & \multicolumn{3}{c}{\textbf{RoboTwin-2.0}} & \multicolumn{3}{c}{\textbf{RoboCasa-GR1}} \\
\cmidrule(lr){3-5} \cmidrule(lr){6-8}
& avg
& Clean $50 \times 50$  & +Random $\times100$ & +Random $\times500$ & $24\times10$ & $24\times100$ & $24\times1000$ \\
\midrule
% \midrule
StarVLA-$\alpha$     & \textbf{98.8} & 50.3 & 78.5 & \textbf{88.2} & 9.8 & 39.4 & 53.8 \\
+ Proprioception            & 98.5 & \textbf{60.8} & \textbf{79.6} & 88.0 & 12.5 & 42.1 & 54.2 \\
+ History frames     & 97.8 & 44.8 & 76.2 & 87.4 & 10.2 & 33.2 & 52.6 \\
+ Delta action            & 98.1 & 48.7 & 77.8 & 85.6 & \textbf{15.8} & \textbf{43.2} & 54.8 \\
+ Relative action           & 98.7 & 51.1 & 77.9 & 87.3 & 13.6 & 40.6 & \textbf{55.5} \\

\bottomrule
\end{tabular}
\end{adjustbox}
\end{table}

\section{All-in-one Evaluation as a Generalist}
\label{sec:all_in_one}

In Sec.~\ref{sec:clean-vla}, we build a clean VLA framework that achieves strong performance across several individual benchmarks. In Sec.~\ref{sec:rethink}, we further rethink several existing techniques and analyze their impact on model training. Hence, after examining the factors that influence training, we move a step further in this section and investigate: \textit{What is an effective evaluation paradigm for assessing whether a model truly possesses generalization ability?}

\mypara{Existing evaluation patterns.} The Embodied AI community shares a unified ambition: to develop a generalist agent that can seamlessly operate across diverse tasks, environments, and robots. In practice, however, the research landscape remains fragmented. Several state-of-the-art systems need to fine-tune their models on benchmark-specific datasets to achieve strong results on individual benchmarks, but their performance drops sharply on others. This leads to a concerning trend in the field: newly proposed policies that excel on one benchmark often suffer sharp performance degradation when transferred to another, making it difficult to demonstrate true generalization ability.

\mypara{All-in-one evaluation as a generalist.} In recent years, large language models (LLMs) have achieved remarkable success, demonstrating generalization capabilities across diverse tasks. A unified evaluation paradigm, which requires a single model to handle multiple benchmarks simultaneously, has driven progress in generalization within the LLM field. This suggests that an appropriate evaluation paradigm can meaningfully shape both model development and the broader direction of the field. Hence, intuitively, Embodied AI should undergo a similar paradigm shift: evaluating a single model across a wide range of diverse benchmarks to ensure that its capabilities are not tied to any specific environment.

\subsection{Task Settings}
In this setting, we utilize all datasets to train a single model jointly and directly evaluate it on multiple benchmarks, without any additional fine-tuning on benchmark-specific datasets. Specifically, we select LIBERO, SimplerEnv, RoboTwin 2.0, and RoboCasa-GR1 as the unified benchmark suite and train the model on the combined training sets of these benchmarks.

\subsection{Experiments}
\mypara{Implementation details.}
We set the learning rate as $1\times10^{-4}$, batchsize as 256 and train on the 5 datasets. In addition, to address the differences in action dimensions across robots, we do not introduce any task-specific design. Instead, we pad the action space of robots with lower degrees of freedom so that all action vectors are uniformly expanded to 32 dimensions in our setting.

\mypara{Baselines.}
To further demonstrate the effectiveness of our method and the proposed setting, we report both specialist results, where models are trained only on individual datasets, and results from the generalist training setting. In addition to comparing with our model, we also evaluate several state-of-the-art methods, such as $\pi_{0.5}$ and GR00T-N1.6.

\mypara{Results.}
As shown in Table~\ref{tab:general-results}, we compare our model trained under the generalist setting, where all datasets are jointly used for training, with specialist models trained on individual benchmarks. Our generalist model consistently achieves sota or competitive performance across most benchmarks. In particular, on the challenging RoboCasa-GR1 benchmark with 24 sub-tasks, our jointly trained model improves performance by 3.5\%. These results suggest that a single model can effectively handle diverse tasks and robot embodiments, supporting the development of more unified evaluation paradigms for embodied AI.

\begin{table}[t]
\centering
\caption{\textbf{Performance comparison between generalist and specialist settings.} Specialist represents multiple models trained separately on each benchmark-specific dataset, while Generalist represents a single model jointly trained across all datasets.
}
\label{tab:general-results}

\begin{adjustbox}{max width=\linewidth}
\begin{tabular}{l l c c c c c c c c c c c c}  % <-- 14 columns
\toprule
\multirow{2}{*}{\textbf{Settings}} & \multirow{2}{*}{\textbf{Method}}
& \multicolumn{5}{c}{LIBERO}
& \multicolumn{3}{c}{SimplerEnv}
& \multicolumn{3}{c}{RoboTwin 2.0}
& \multirow{1}{*}{RoboCasa-GR1} \\
\cmidrule(lr){3-7} \cmidrule(lr){8-10} \cmidrule(lr){11-13} \cmidrule(lr){14-14}
& & Spatial & Object & Goal & Long & avg
& WidowX & Google VA & Google VM
& clean & clean$^*$ & random$^*$
& (avg of 24 tasks) \\
\midrule

\multirow{5}{*}{\textbf{Specialist}}
& $\pi_{0.5}$   & 98.8 & 98.2 & 98.0 & 92.4 & 96.9 & 46.9  & 68.4 & 72.7 &  \textbf{60.2} & 82.7 & 76.8 & 37.0 \\
& GR00T-N1.6    & 97.5 & 98.5 & 97.5 & 94.4 & 94.1 & \textbf{67.8} & 41.5 & 35.2 & --   & --   & --   & 47.6 \\
\cmidrule(lr){2-14}
& StarVLA-$\alpha$-$\pi$     & 98.0 & 99.2 & 98.2 & 93.6 & 98.1 & 65.9 & \textbf{72.8} & \textbf{76.6} & 50.8 & 88.1 & \textbf{88.8} & 48.9 \\
& StarVLA-$\alpha$-GR00T  & 98.9 & 99.6 & 98.4 & \textbf{95.3} & 98.7 & 65.3 & 70.7 & 75.3 & 48.8 & 88.0 & 88.5 & 52.8 \\
& StarVLA-$\alpha$    & \textbf{99.0} & \textbf{99.8} & 98.5 & 94.1 & \textbf{98.8} & 64.6 & 70.2 & 76.0 & 53.4 & 88.2 & 88.3 & 53.8 \\
\midrule
\textbf{Generalist}
& StarVLA-$\alpha$  & 98.7 & 99.7 & \textbf{98.6} & 94.2 & 97.8 & 65.2 & 69.8 & 74.3 & --   & \textbf{88.7} & 87.8 & \textbf{57.3} \\
\bottomrule
\end{tabular}
\end{adjustbox}
\end{table}

\subsection{Discussion and Analysis}
Our method is simple: it directly pads all actions to the same dimension and uses Qwen3-VL as the pretrained model, yet achieves strong performance. Therefore, in this section, we discuss and analyze \textit{what the most critical factor is in this generalist setting} and \textit{why such a simple method performs so strongly}. We examine this question from multiple perspectives, including action processing, model size, model initialization, and the impact of batch size.

\begin{figure}
    \centering
    \includegraphics[width=1\linewidth]{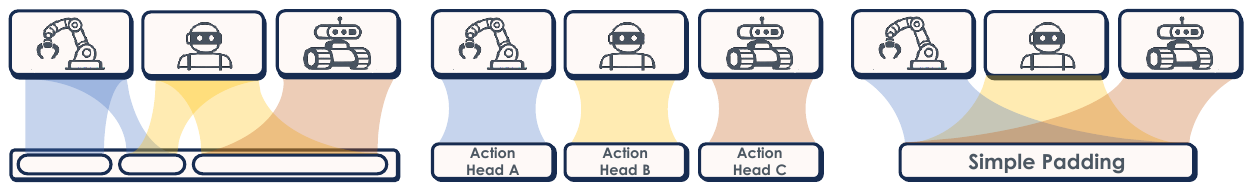}
    \caption{\textbf{Comparison of action parameterization for multiple embodiments.} Left: RDT Action. Middle: Multi-Action Head. Right: Simple Padding strategy.}
    \label{fig:cross-embodiment}
\end{figure}

\mypara{Do we truly need specific action designs for each embodiment?}
Previous studies, such as ABot-VLA~\cite{yang2026abot} and LingBot-VLA~\cite{wu2026pragmatic}, have proposed complex, robot-specific solutions, including unified action spaces and multi-action heads tailored to each robotic embodiment. However, modern vision–language models (VLMs) possess sufficient intelligence and parameter capacity to handle diverse tasks. Therefore, can we instead adopt a simple padding strategy and allow the VLA model itself to recognize and manage tasks across multiple embodiments?
As shown in Table~\ref{tab:pading-method}, we compare simple padding strategy with RDT Action and the Multi-Action Head (Fig.~\ref{fig:cross-embodiment}). Our approach achieves comparable performance on LIBERO and RoboTwin 2.0, while improving results on Google Robot VM and RoboCasa-GR1 by 2.9\% and 4.8\%, respectively. These results suggest that complex specialist designs may be unnecessary for challenging cross-embodiment tasks.

\begin{table}[t]
\centering
\caption{{\textbf{Performance comparison of multi-embodiment action parameterization}. }}
\label{tab:pading-method}
\begin{adjustbox}{max width=\linewidth}
\begin{tabular}{l c c c c c c c}
\toprule
\multirow{2}{*}{\textbf{Method}} & \multicolumn{1}{c}{\textbf{LIBERO}}
& \multicolumn{3}{c}{\textbf{SimplerEnv}}
& \multicolumn{2}{c}{\textbf{RoboTwin 2.0}}
& \multicolumn{1}{c}{\textbf{RoboCasa-GR1}} \\
\cmidrule(lr){2-2}\cmidrule(lr){3-5}\cmidrule(lr){6-7}\cmidrule(lr){8-8}
& Avg
& \shortstack{WidowX}
& \shortstack{Google VA}
& \shortstack{Google VM}
& clean\*
& random\*
& avg \\
\midrule
RDT action~\cite{liu2024rdt}          & 97.2 & 63.9 & 68.2 & 71.4 & 87.2 & 86.6   & 52.3 \\
Multi Action Header~\cite{hpt} & 97.2 & 60.6 & 66.3 & 67.8 & 85.6 & 86.1   & 53.5 \\
Simple Padding~\cite{pi_0}      & \textbf{97.8} & \textbf{65.2} & \textbf{69.8} & \textbf{74.3} & \textbf{88.7} & \textbf{87.8}   & \textbf{57.3} \\

\bottomrule
\end{tabular}
\end{adjustbox}
% \vspace{-1em}
\end{table}

% \vspace{-1em}
\begin{figure}
    \centering
    \includegraphics[width=0.95\linewidth]{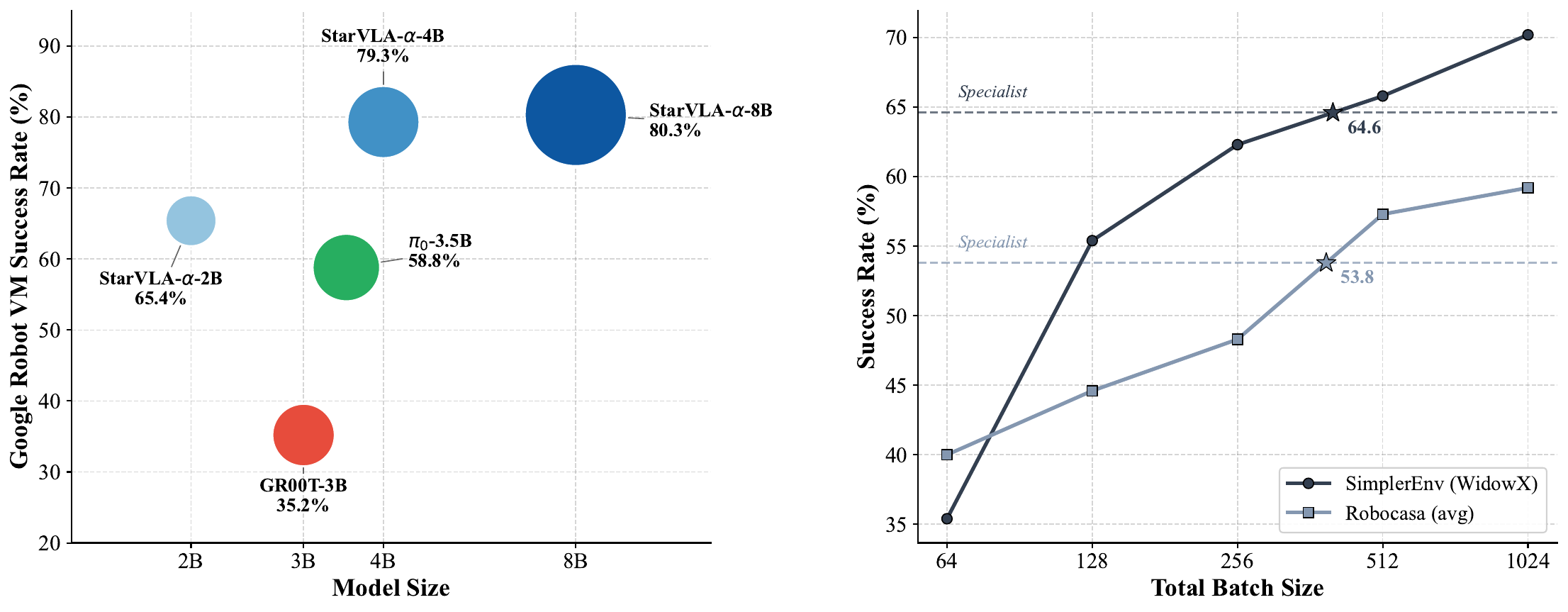}
    \caption{\textbf{Scaling trends in VLA training.} Left: performance as a function of model size. Right: performance as a function of total batch size.}
    \label{fig:modelsize-batchsize}
\end{figure}

% \vspace{-2em}
\mypara{What is the influence of model size?}
To further investigate the impact of model size on VLA performance in this general all-in-one setting, we evaluate three pretrained Qwen3-VL models (2B, 4B, and 8B) under the same experimental setup. As shown in Fig.~\ref{fig:modelsize-batchsize}, the 4B model achieves significant performance improvements on Simpler compared with the 2B model. Additional results in Appendix~\ref{sec:more_abl} further show that this improvement generalizes beyond a single benchmark, yielding gains of 18.1\% on WidowX and 6.6\% on RoboCasa-GR1. However, compared with the 4B model, the 8B model does not demonstrate substantial additional improvements, with gains remaining within 1\%. These results suggest that, under the current training scale and scenarios, the model size should not be too small, but a 4B parameter scale is sufficient.

\begin{table}[t]
\centering
\small
\caption{\textbf{Real-world evaluation as a Generalist in RoboChallenge}. SR represents success rate, and score represents progress score.}
\label{tab:arx5_cleanvla}
\renewcommand{\arraystretch}{1.15}

\resizebox{0.7\linewidth}{!}{%
\begin{tabular}{l l|cc|cc|cc}
\toprule
\multirow{2}{*}{Robot} & \multirow{2}{*}{Task}
& \multicolumn{2}{c|}{\makecell{\textbf{StarVLA-$\boldsymbol{\alpha}$}}}
& \multicolumn{2}{c|}{\makecell{\textbf{$\pi_{0.5}$}}}
& \multicolumn{2}{c}{\makecell{\textbf{$\pi_{0}$}}} \\
\cmidrule(lr){3-4}\cmidrule(lr){5-6}\cmidrule(lr){7-8}
& & \cellcolor{yellow!25}\textbf{SR} & \cellcolor{yellow!25}\textbf{score}
  & \cellcolor{yellow!25}\textbf{SR} & \cellcolor{yellow!25}\textbf{score}
  & \cellcolor{yellow!25}\textbf{SR} & \cellcolor{yellow!25}\textbf{score} \\
\midrule

\multirow{12}{*}{ARX5}
& arrange flowers               & 40.0  & 66.5 & 0.0  & 30.5  & 0.0  & 13.5 \\
& arrange paper cups            & 20.0  & 63.0 & 0.0  & 31.0  & 0.0  & 15.0 \\
& fold dishcloth                & 0.0   & 3.5  & 0.0  & 0.0   & 0.0  & 0.0  \\
& open the drawer               & 20.0  & 60.0 & 50.0 & 80.0  & 0.0  & 20.0 \\
& place shoes on rack           & 50.0  & 70.0 & 0.0  & 20.0  & 0.0  & 16.5 \\
& put cup on coaster            & 100.0 & 98.0 & 70.0 & 63.0  & 0.0  & 0.0  \\
& search green boxes            & 60.0  & 58.5 & 0.0  & 3.0   & 0.0  & 0.0  \\
& sort electronic products      & 20.0  & 39.4 & 0.0  & 22.5  & 0.0  & 22.5 \\
& turn on light switch          & 50.0  & 59.0 & 10.0 & 25.0  & 20.0 & 29.0 \\
& water potted plant            & 10.0  & 32.0 & 0.0  & 0.0   & 0.0  & 0.0  \\
& wipe the table                & 0.0   & 44.5 & 10.0 & 28.0  & 0.0  & 29.0 \\
\cmidrule(lr){2-8}
& \textbf{Avg.}                 & \textbf{33.6} & \textbf{54.5} & \textbf{12.7} & \textbf{27.6} & \textbf{3.6} & \textbf{14.7} \\
\bottomrule
\end{tabular}%
}
\end{table}

\mypara{Does batchsize matter in the generalist settings? Definitely Yes!}
As shown in Fig.~\ref{fig:modelsize-batchsize}, with additional results provided in Appendix~\ref{sec:more_abl}, we evaluate batch sizes of 64, 128, 256, 512, and 1024 under the same total training scale. Performance improves consistently as the batch size increases. With a batch size of 512, the model already achieves strong performance, e.g., 57.3 on RoboCasa-GR1 and 57.2 on RoboTwin-Clean.
These results indicate that a larger batch size, which ensures sufficient diversity during training, is a key factor. It helps prevent the model from becoming trapped in local minima, thereby enabling better generalization.

\section{Real-World Experiments}
\label{sec:eval_real}

In this section, we validate that our minimalist framework remains competitive in physical robot experiments. We evaluate our model on the public real-world benchmark \textbf{RoboChallenge}~\cite{robochallenge}, which provides standardized tasks for direct comparison with existing models. Additional real-world OOD experiments are provided in the Appendix.

\mypara{Benchmark.} RoboChallenge is a large-scale real-robot evaluation platform designed to assess learned robotic control policies on physical hardware in a standardized and reproducible manner. We evaluate on the RoboChallenge suite, which contains several tabletop manipulation tasks (e.g., object reorientation, insertion, and multi-stage operations). Following the benchmark protocol, each task is executed multiple times to reduce stochasticity in real-robot trials, and performance is measured by the average success rate under predefined task success criteria. Additional implementation details are provided in the Appendix.

\mypara{Results.} 
As shown in Table~\ref{tab:arx5_cleanvla}, on the ARX5 robot, we execute the standard set of 11 tasks. The results show that our StarVLA-$\alpha$ achieves a success rate of 33.6 and a progress score of 54.5, which significantly surpass the 12.7 and 27.6 achieved by $\pi_{0.5}$, respectively. These results demonstrate the effectiveness of our StarVLA-$\alpha$ in real-world settings.

\section{Related Works}
\label{sec:rel}

\noindent{\textbf{Vision-language-action (VLA) models}}. 
The rapid progress of Large Vision-Language Models (VLMs)~\cite{beyer2024paligemma, llava, qwen2vl} has catalyzed a paradigm shift toward end-to-end Vision-Language-Action (VLA) policies~\cite{RT-1, RT-2}. 
Building on this foundation, recent work has rapidly expanded the VLA paradigm, exploring diverse architectural designs, including decoupled vision encoder–LLM pipelines~\cite{llava, openvla}, native multimodal models~\cite{qwen3vl}, and specialized action decoding mechanisms~\cite{pi_05, bjorck2025gr00t}. 
Meanwhile, training strategies vary substantially across works, spanning robotic datasets~\cite{zheng2025x, zheng2025universalactionsenhancedembodied}, human video demonstrations~\cite{lapa, li2024decisionnce}, and web data co-training~\cite{pi_05}. 
Alongside a growing number of architectural variants~\cite{reconvla, cogact, spatialvla} and training recipes~\cite{openvla-oft, wang2025vlaadapter, zheng2025x}, these design choices introduce significant heterogeneity across VLA systems, making it difficult to attribute performance improvements to specific algorithmic innovations.

\noindent{\textbf{Robotic data engineering and action parameterization}}. 
 Datasets for robot learning~\cite{open_x_embodiment, khazatsky2024droid} require extensive preprocessing to reconcile differences in control frequencies, camera viewpoints, and action formats~\cite{robomind}. Diverse action parameterization strategies, ranging from discretized token prediction~\cite{RT-1, RT-2, openvla} and continuous autoregressive control~\cite{openvla-oft} to action chunking~\cite{aloha} and diffusion-based policies operating~\cite{pi_05, chi2024diffusionpolicyvisuomotorpolicy}, have been extensively explored across different models. In addition, various data processing strategies have been shown to affect downstream performance~\cite{wang2025vlaadapter}, including normalization schemes~\cite{openvla}, proprioceptive state conditioning~\cite{reuss2025flower}, and cross-embodiment padding mechanisms~\cite{octo, liu2024rdt}. The reliance on heterogeneous data pipelines tightly couples algorithmic design with engineering choices, obscuring the true sources of performance gains.

\section{Conclusion}

We introduced \textbf{StarVLA-$\boldsymbol{\alpha}$}, a simple VLA baseline combining a strong VLM backbone with a lightweight MLP action head and minimal data processing, which achieves strong performance across multiple benchmarks and real-world robotic tasks. Controlled experiments with StarVLA-$\alpha$ show that many complex techniques, i.e., sophisticated action header design, heavy data engineering, or task-specific pretraining, are not strictly necessary for generalist robot development. This simplified design reduces architectural complexity, minimizes data engineering, and provides a reproducible and generalizable framework for future VLA research.

\section*{Acknowledgements}
This work was supported in part by the Research Grants Council under the Areas of Excellence scheme grant AoE/E-601/22-R and by ACCESS - AI Chip Center for Emerging Smart Systems, supported by the InnoHK initiative of the Innovation and Technology Commission of the Hong Kong Special Administrative Region Government.

{\small
\bibliographystyle{apalike}  % Or any other style that suits your requirements
\bibliography{egbib}
}

\newpage
\appendix

\appendix

\begin{center}\large\bf
Supplementary Material for ``StarVLA-$\boldsymbol{\alpha}$: Reducing Complexity in Vision-Language-Action Systems''
\end{center}

\vspace{0.5em}
The supplementary material is organized as follows.

\begin{enumerate}[leftmargin=0.22in]
    \item \textbf{Related works.} Related works on VLA models, robotic data engineering, and action parameterization are described in Sec.~\ref{sec:related_works}.

    \item \textbf{Benchmark details.} Detailed descriptions of all benchmarks, including LIBERO, SimplerEnv, RoboTwin 2.0, RoboCasa-GR1, and RoboChallenge, are described in Sec.~\ref{sec:benchmark_detail}.

    \item \textbf{Training details.} Default training setup, optimization hyperparameters, compute resources, and architecture details are described in Sec.~\ref{sec:training_detail}.

    \item \textbf{More ablation studies.} Additional ablations on model initialization, model size, and batch size in the all-in-one setting are described in Sec.~\ref{sec:more_abl}.

    \item \textbf{Large-scale real-world evaluations on RoboChallenge.} Large-scale real-world evaluation results on the RoboChallenge benchmark~\cite{robochallenge} across multiple robot embodiments as a Generalist are described in Sec.~\ref{sec:robochallenge}. 
    % \yilun{TODO: highlight Generalist means cross-embodiment.}

    \item \textbf{Real-world OOD experiments.} Experimental setup and results for real-world out-of-distribution evaluation are described in Sec.~\ref{sec:real_ood}.

    \item \textbf{Detailed benchmarks results.} Full benchmark results and supplementary quantitative comparisons are described in Sec.~\ref{sec:full_results}.

    \item \textbf{Qualitative results across simulation benchmarks.} Visualizations of simulation benchmarks, RoboChallenge, and real-world deployment settings are described in Sec.~\ref{sec:vis_all_benchmarks}.

    \item \textbf{Robustness evaluation on LIBERO-Plus.} Additional robustness evaluation results on the LIBERO-Plus benchmark are described in Sec.~\ref{sec:libero_plus}.
\end{enumerate}

\section{Related Works} \label{sec:related_works}

\noindent{\textbf{Vision-language-action (VLA) models}}. The rapid advancement of Large Vision-Language Models(VLMs)~\cite{beyer2024paligemma, llava, qwen2vl} has fundamentally reshaped the development of robotics models, driving a paradigm shift toward end-to-end Vision-Language-Action(VLA) frameworks. By directly mapping multimodal observations to deployable control signals, pioneering works like RT-series~\cite{RT-1, RT-2} demonstrated the viability of leveraging VLM reasoning for generalist embodied agents. Building upon this foundation, the community has witnessed a surge of VLA methodologies. Initiatives like Octo~\cite{octo} and OpenVLA~\cite{openvla} explored diverse backbone with specific injection methods for action, while recent advancement like $\pi$-series~\cite{pi_0, pi_05} and GR00T-series~\cite{bjorck2025gr00t} introduced specialized action decoding mechanism and larger-scale robotics pretraining. However, the rapid iteration within the field introduce massive heterogeneous structural designs~\cite{reconvla, cogact, spatialvla} and disparate training recipes~\cite{openvla-oft, wang2025vlaadapter, zheng2025x}. For example, the choice of VLM backbone vary drastically across models, ranging from decoupled vision-encoder-plus-LLM pipelines~\cite{llava, openvla} to natively multimodal architectures~\cite{qwen3vl}, while pre-training recipe diverge significantly among cross-embodiment~\cite{zheng2025x, zheng2025universalactionsenhancedembodied}, human video~\cite{lapa, li2024decisionnce} and even vision-language data co-training~\cite{pi_05}. Furthermore, many approaches rely heavily on idiosyncratic engineering practice for specific evaluation recipe~\cite{openvla-oft,wang2025vlaadapter}, resulting in a highly fragmented methodological landscape and unreliable evaluation. In this work, we build a clean and neat framework to abstract away this structural complexity, establishing a rigorously controlled baseline to isolate the true drivers of VLA performance.

\noindent{\textbf{Robotic data engineering and action parameterization}}. The landscape of robotic learning has been significantly propelled by advancing data engineering techniques. To effectively harness heterogeneous datasets~\cite{open_x_embodiment, khazatsky2024droid} and promote learning efficiency, the community has introduced a variety of specialized techniques. A central challenge lies in transforming incompatible control frequencies, camera viewpoints, and action representations into formats compatible with VLM-based policies~\cite{robomind}, which often requires substantial dataset-specific preprocessing. At the same time, the design of action parameterization remains an actively debated topic~\cite{feng2026demystifying}. Implementations span a broad spectrum: from discretizing continuous control signals into language tokens via uniform binning~\cite{RT-1, RT-2, openvla}, to continuous autoregressive prediction~\cite{openvla-oft}, action chunking~\cite{aloha}, and high-frequency diffusion processes~\cite{pi_05,chi2024diffusionpolicyvisuomotorpolicy}. Beyond action modeling, auxiliary data processing choices, ranging from dataset-specific normalization schemes~\cite{openvla} and conditional injection of proprioceptive states~\cite{reuss2025flower} to complex padding mechanisms for cross-embodiment alignment~\cite{octo, liu2024rdt}, have been successively demonstrated across various studies to substantially impact downstream performance~\cite{wang2025vlaadapter}. This deep reliance on disparate data pipelines and action modeling deeply entangles algorithmic innovations with empirical tuning, obscuring whether performance gains originate from core architectural advancements or merely optimized recipes. In this work, we systematically disentangle these confounding factors by establishing a clean, unified baseline that isolates and rigorously evaluates the true impact of these individual engineering choices under strictly controlled conditions.

\section{Benchmark Details}
\label{sec:benchmark_detail}

We evaluate StarVLA-$\alpha$ on a diverse set of benchmarks that cover complementary aspects of robotic manipulation, including compositional multi-task learning~\cite{Libero, libero_plus}, simulated evaluation of real-world policies~\cite{simpleenv}, dual-arm coordination~\cite{chen2025robotwin2}, humanoid tabletop manipulation~\cite{robocasa}, and standardized real-robot testing~\cite{robochallenge}. Together, these benchmarks span different embodiments, task structures, and evaluation protocols, providing a broad testbed for studying both specialization and generalization in VLA systems.

\begin{itemize}[leftmargin=0.2in]

    \item \textbf{LIBERO:}
    LIBERO is a widely used benchmark for language-conditioned robot manipulation and lifelong robot learning~\cite{Libero}. The benchmark contains \textit{130 manipulation tasks} organized into four task suites, namely \textit{Spatial}, \textit{Object}, \textit{Goal}, and \textit{Long}. The first three suites are designed to isolate transfer under controlled shifts in spatial relations, object identities, and task goals, while LIBERO-Long contains a larger set of manipulation tasks with more entangled variations. In the literature, a common training protocol uses 50 expert demonstrations per task, yielding approximately 6.5K trajectories across the full benchmark. LIBERO provides a standardized evaluation protocol and has become a common testbed for studying instruction following, compositional generalization, and multi-task policy learning in language-conditioned manipulation settings. We additionally evaluate our model on LIBERO-Plus~\cite{libero_plus}, which is a robustness-oriented benchmark built on top of LIBERO for systematically evaluating VLA policies under controlled distribution shifts.

    \item \textbf{SimplerEnv:}
    SimplerEnv is a simulation framework designed for evaluating real-world manipulation policies in simulation~\cite{simpleenv}. Rather than focusing on policy learning in simulation, it provides a standardized and scalable proxy for real-world evaluation, and its results have been shown to correlate with physical robot performance. The benchmark includes simulated environments corresponding to common real-robot setups, in particular the Google Robot environments used in the RT-series and the WidowX / BridgeData V2 setting. As a result, SimplerEnv is widely used to assess whether a policy trained on real-world robot data can generalize to standardized evaluation scenarios without requiring direct real-robot testing for every experiment.

    \item \textbf{RoboTwin 2.0:}
    RoboTwin 2.0 is a large-scale benchmark for \textit{bimanual robotic manipulation} that focuses on dual-arm coordination across diverse interaction scenarios~\cite{chen2025robotwin2}. We use 50 clean data per task for standard clean evaluation, and followed the multi-task training protocol uses 50 clean demonstrations per task together with 500 randomized demonstrations per task, resulting in approximately 550 trajectories per task and 27.5K trajectories over all 50 tasks. The randomized trajectories are generated through structured domain randomization, typically including factors such as cluttered scenes, background variation, table-height perturbation, lighting changes, and other environmental variations. This benchmark therefore provides a challenging testbed for evaluating fine-grained bimanual coordination as well as robustness to environmental diversity.

    \item \textbf{RoboCasa-GR1:}
    RoboCasa-GR1 is a tabletop manipulation benchmark built on top of the RoboCasa simulation framework and is commonly used for evaluating humanoid-style manipulation policies~\cite{robocasa,bjorck2025gr00t}. Compared with standard single-arm tabletop benchmarks, it introduces a more challenging embodiment together with household interaction scenarios involving articulated objects and multi-stage manipulation. The benchmark contains 24 tabletop tasks, and the associated data release commonly used in recent work provides 1000 demonstrations per task, resulting in around 24K trajectories in total. RoboCasa-GR1 is therefore a useful benchmark for studying cross-embodiment transfer, long-horizon tabletop manipulation, and humanoid-oriented visuomotor control.

    \item \textbf{RoboChallenge:}
    RoboChallenge is a large-scale real-robot evaluation platform for benchmarking embodied control policies in the physical world~\cite{robochallenge}. Its initial benchmark suite, \textbf{Table-30}, contains 30 real-world tabletop manipulation tasks, and the platform report describes an initial deployment with \textbf{10 hosted machines}. Unlike simulation benchmarks, RoboChallenge evaluates policies directly under real sensing noise, actuation uncertainty, and physical environment variability. It therefore serves as an important testbed for validating whether strong simulation performance can transfer to standardized real-world robotic deployment.

\end{itemize}

\section{Training Details}
\label{sec:training_detail}

\mypara{Default setup.} Unless otherwise specified, we initialize the VLM backbone from the publicly available Qwen3-VL-4B checkpoint, while the action heads are randomly initialized. All models are trained in a single stage directly on the target benchmark data without any action-specific pretraining.

\mypara{Training paradigms.} We study both \textit{\textbf{Specialist}} and \textit{\textbf{Generalist}} training. A \textit{Specialist} model is trained using data from a single embodiment only. In contrast, a \textit{Generalist} model is trained jointly on merged data from multiple embodiments and benchmark suites. In this paper, all benchmarks jointly training correspond to the \textit{Generalist} setting.

\mypara{Optimization.} We use different learning rates for the VLM backbone and the action head: $1\times10^{-5}$ for the backbone and $1\times10^{-4}$ for the action head, with a cosine learning rate schedule. All models are trained for a maximum of 100k steps with a per-GPU batch size of 16.

\mypara{Computation resources.} Our training setup scales with dataset size while keeping all other hyperparameters benchmark-agnostic. Specifically, shown in Table~\ref{tab:compute_resources}:

\begin{table}[h]
\centering
\caption{ \textbf{Computation resources for each benchmark suite.} }
\label{tab:compute_resources}
\small
\setlength{\tabcolsep}{6pt}
\begin{tabular}{l c l c}
\toprule
\textbf{Training Data} & \textbf{\#GPUs} & \textbf{Training Data} & \textbf{\#GPUs} \\
\midrule
LIBERO & $8\times$ A100 &  SimplerEnv & $16\times$ A100 \\
RoboCasa-GR1 & $16\times$ A100 & RoboTwin-Clean & $16\times$ A100 \\
RoboTwin-Clean + Rand. & $48\times$ A100 & RoboChallenge Table 30 & $32\times$ A100 \\
Real-World OOD  & $16\times$ A100 & 
\multicolumn{1}{l}{All benchmarks jointly} & $64\times$ A100 \\
\bottomrule
\end{tabular}
\end{table}

\mypara{Architecture details.} For implementation specifics of different action heads (FAST, OFT, PI, GR00T), we refer readers to our code at 
\url{https://github.com/starVLA/starVLA/tree/starVLA/starVLA/model/modules/action_model}.

\section{More Ablation Studies}
\label{sec:more_abl}

In addition to the analyses in Sec.~\ref{sec:all_in_one}, we further study several practical factors that may influence generalist VLA training: {model initialization}, {model size}, and {batch size}. Unlike the main ablations in Sec.~\ref{sec:rethink}, these experiments focus on the all-in-one training setting and aim to better understand which factors are most critical for achieving strong cross-benchmark performance under a unified training pipeline.

\subsection{Effect of Model Initialization}
\label{sec:abl_model_init_general}

\mypara{Motivation.}
A key question in the generalist setting is whether the gain mainly comes from the unified training recipe itself, or depends critically on the backbone initialization. Since our framework is built on top of a pretrained VLM, it is important to quantify the role of initialization quality.

\mypara{Experimental setup.}
We keep the all-in-one training recipe unchanged and vary only the backbone initialization. Specifically, we compare random initialization, Qwen2.5-VL, and Qwen3-VL, while using the same action head and training pipeline in all cases.

\begin{table}[h]
\centering
\caption{\textbf{Performance comparison across different VLM model initialization.}}
\label{tab:model_init}
\begin{adjustbox}{max width=\linewidth}
\begin{tabular}{l c c c c c c c}
\toprule
\multirow{2}{*}{\textbf{VLM Initial}} & \multicolumn{1}{c}{\textbf{LIBERO}}
& \multicolumn{3}{c}{\textbf{SimplerEnv}}
& \multicolumn{2}{c}{\textbf{RoboTwin 2.0}}
& \multicolumn{1}{c}{\textbf{RoboCasa-GR1}} \\
\cmidrule(lr){2-2}\cmidrule(lr){3-5}\cmidrule(lr){6-7}\cmidrule(lr){8-8}
& Avg
& \shortstack{WidowX}
& \shortstack{Google VA}
& \shortstack{Google VM}
& clean\*
& random\*
& Avg \\
\midrule
Random            & 77.5          & 24.8          & 45.4          & 52.8          & 65.6          & 59.8 & 28.8 \\
Florence-2          & 93.2 &   53.4 & 63.6 & 65.7 & 77.8 & 79.1 & 39.2 \\
Qwen2.5-VL        & 95.5          & 65.6          & 70.7          & 77.1          & 87.2          & 85.6 & 53.6 \\
Qwen3-VL          & {97.8} & {70.2} & {73.8} & {79.3} & \textbf{88.7} & {87.8} & \textbf{57.3} \\
Qwen3.5          &  \textbf{98.2} & \textbf{71.3} & \textbf{76.8} & \textbf{80.0} & {88.3} & \textbf{88.4} & 56.1 \\
\bottomrule
\end{tabular}
\end{adjustbox}
\end{table}

\mypara{Results.} 
Table~\ref{tab:model_init} shows that initialization quality plays a crucial role in generalist policy training. Random initialization performs substantially worse across all benchmarks, indicating that unified training alone is insufficient to learn a strong generalist policy from scratch. In contrast, initializing from pretrained VLMs consistently improves performance. Among the pretrained backbones, stronger models generally lead to better results. Qwen2.5-VL already provides a large improvement over Florence-2, and Qwen3-VL further improves performance across most benchmarks. Using the even stronger Qwen3.5 backbone yields the best results on LIBERO and SimplerEnv and achieves the highest average performance on RoboTwin 2.0, demonstrating that improved multimodal priors can further enhance downstream robot control.

\mypara{Takeaway.}
These results indicate that \textbf{strong VLM initialization is a key ingredient for generalist VLA training}. The backbone is not merely a starting point: better pretrained multimodal representations translate directly into stronger cross-benchmark generalization.

\subsection{Effect of Model Size}
\label{sec:abl_model_size_general}

\mypara{Motivation.}
Besides initialization, model capacity may also affect how well a single policy absorbs heterogeneous data from multiple benchmarks and embodiments. We therefore study whether scaling the backbone improves performance in the generalist setting.

\mypara{Experimental setup.}
We evaluate three Qwen3-VL model sizes, 2B, 4B, and 8B, under the same all-in-one training setup. All other settings, including the action head, optimizer, and unified action padding strategy, remain unchanged.

\begin{table}[h]
\centering
\caption{\textbf{Performance across VLA model sizes under \textit{Generalist} setting.}}
\label{tab:geneal_model_size}
\begin{adjustbox}{max width=\linewidth}
\begin{tabular}{l c c c c c c c}
\toprule
\multirow{2}{*}{\textbf{Method}} & \multicolumn{1}{c}{\textbf{LIBERO}}
& \multicolumn{3}{c}{\textbf{SimplerEnv}}
& \multicolumn{2}{c}{\textbf{RoboTwin 2.0}}
& \multicolumn{1}{c}{\textbf{RoboCasa-GR1}} \\
\cmidrule(lr){2-2}\cmidrule(lr){3-5}\cmidrule(lr){6-7}\cmidrule(lr){8-8}
& Avg
& \shortstack{WidowX}
& \shortstack{Google VA}
& \shortstack{Google VM}
& clean*
& random*
& Avg \\
\midrule
2B & 97.8 & 52.1 & 61.5 & 65.4 & 76.8 & 79.1 & 50.7 \\
4B & {97.8} & 70.2 & \textbf{73.8} & 79.3 & \textbf{88.7} & 87.8 & 57.3 \\
8B & \textbf{ 98.2 }& \textbf{71.5} & 72.9 & \textbf{80.3} & 88.6 & 88.8 & \textbf{58.2} \\
\bottomrule
\end{tabular}
\end{adjustbox}
\end{table}

\mypara{Results.}
As shown in Table~\ref{tab:geneal_model_size}, increasing model size from 2B to 4B brings clear and consistent gains, especially on the more challenging benchmarks. This suggests that insufficient capacity can limit unified multi-benchmark learning even when the training recipe is fixed. However, the improvement from 4B to 8B is much smaller and less consistent, indicating a clear diminishing-return trend once the backbone reaches moderate scale.

\mypara{Takeaway.}
These results suggest that \textbf{model size should not be too small in the generalist setting, but scaling beyond a moderate size is not the dominant factor}. In our setup, 4B already captures most of the achievable gains and provides a favorable trade-off between capacity and efficiency.

\subsection{Effect of Batch Size}
\label{sec:abl_batchsize_general}

\mypara{Motivation.}
In the all-in-one setting, each batch may contain samples from different tasks, robots, and environments. As a result, batch size directly affects how much diversity the model observes at each optimization step, which may be particularly important for cross-benchmark generalization.

\mypara{Experimental setup.}
We vary the total batch size from 64 to 1024 while keeping the rest of the training setup unchanged. The model architecture and unified action representation remain the same, so the effect can be attributed to batch size alone.

\begin{table}[h]
\centering
\caption{\textbf{Performance comparison under different batch sizes.}}
\label{tab:general_batchsize}
\begin{adjustbox}{max width=\linewidth}
\begin{tabular}{l c c c c c c c}
\toprule
\multirow{2}{*}{\textbf{ Batch Size}} & \multicolumn{1}{c}{\textbf{LIBERO}}
& \multicolumn{3}{c}{\textbf{SimplerEnv}}
& \multicolumn{2}{c}{\textbf{RoboTwin 2.0}}
& \multicolumn{1}{c}{\textbf{RoboCasa-GR1}} \\
\cmidrule(lr){2-2}\cmidrule(lr){3-5}\cmidrule(lr){6-7}\cmidrule(lr){8-8}
& Avg
& \shortstack{WidowX}
& \shortstack{Google VA}
& \shortstack{Google VM}
& clean\*
& random\*
& Avg \\
\midrule
64   & 95.8 & 35.4 & 63.2 & 76.8 & 80.4 & 81.3 & 40.0 \\
128  & 97.4 & 55.4 & 63.8 & 77.2 & 80.8 & 83.8 & 44.6 \\
256  & 97.9 & 62.3 & 70.6 & 78.3 & 86.4 & 86.5 & 48.3 \\
512  & 98.1 & 65.8 & 70.7 & 79.7 & \textbf{88.8} & 88.7 & 57.3 \\
1024 & \textbf{98.8} & \textbf{70.2} & \textbf{71.3} & \textbf{80.1} & \textbf{88.8} & \textbf{89.2} & \textbf{59.2} \\
\bottomrule
\end{tabular}
\end{adjustbox}
\end{table}

\mypara{Results.}
Table~\ref{tab:general_batchsize} shows a clear and consistent benefit from larger batch sizes. Performance improves steadily as the batch size increases, with especially pronounced gains on more challenging benchmarks such as SimplerEnv, RoboTwin 2.0, and RoboCasa-GR1. This trend suggests that, in the unified setting, exposing the model to more diverse supervision within each step is crucial for stable optimization and stronger generalization.

\mypara{Takeaway.}
These results indicate that \textbf{batch size is one of the most important optimization factors in generalist VLA training}. Its impact is broader and more consistent than that of model scaling, highlighting the importance of diverse gradient signals in all-in-one training.

\section{RoboChallenge}
\label{sec:robochallenge}

To further evaluate the real-world performance of StarVLA-$\alpha$, we report results on the RoboChallenge Table-30 benchmark. RoboChallenge is a standardized real-robot benchmark that covers a broad range of household manipulation tasks across multiple robot embodiments. In our evaluation, the benchmark includes four representative platforms, namely UR5, Franka, ARX5, and ALOHA, and a diverse set of tasks with different levels of difficulty, including short-horizon pick-and-place, precise object manipulation, and long-horizon multi-step interaction. This benchmark therefore provides a strong testbed for measuring not only raw task success, but also embodiment-level transfer and robustness across varied real-world manipulation settings.

Table~\ref{tab:full_roboca_cleanvla} summarizes the full evaluation results across all four robot platforms. Each task is evaluated with two complementary metrics: success rate (SR) and progress score. The success rate measures whether the full task is completed successfully, while the progress score captures partial completion and thus provides a finer-grained view of policy behavior on long-horizon or difficult tasks where binary success alone may hide meaningful differences.

Overall, StarVLA-$\alpha$ consistently outperforms $\pi_{0.5}$ and $\pi_{0}$ on most platforms and task groups in terms of both success rate and progress score. The gains are especially clear on challenging tasks that require accurate grounding, sequential reasoning, or stable long-horizon control, showing that StarVLA-$\alpha$ is not only stronger at completing tasks end-to-end, but also more reliable in making steady progress when full completion is difficult. These results validate the effectiveness of StarVLA-$\alpha$ in real-world manipulation and show that even a simple unified framework can generalize well across different robot embodiments and diverse task distributions.

\clearpage
\clearpage

\section{Real-world OOD Experiments}
\label{sec:real_ood}

To further assess the robustness of StarVLA-$\alpha$ in practical deployment, we conduct a set of real-world out-of-distribution (OOD) experiments using a physical robot. Compared with the standardized real-world benchmark results in Sec.~\ref{sec:eval_real}, these experiments are designed to explicitly test whether StarVLA-$\alpha$ can generalize under realistic distribution shifts, including \textbf{novel objects}, \textbf{unseen colors}, \textbf{shifted object positions}, and \textbf{unseen spatial coordinates}. The goal is to evaluate whether the same simple StarVLA-$\alpha$ framework remains reliable in real-world instruction-following tasks beyond benchmark-specific settings.

\mypara{Experimental setup.}
We use a stationary, table-mounted \textbf{Franka Research 3} robot arm for all real-world experiments. The observation consists of two RGB images: one from a fixed third-person camera and the other from a wrist-mounted first-person camera. Both images are resized to $224\times224$ before being fed into the model. We consider three representative real-world manipulation tasks that probe different aspects of OOD generalization. The first is a \textbf{waste-sorting categorization} task, where the robot must place objects into the correct bin according to semantic category; the OOD setting contains \emph{novel objects}. The second is a \textbf{pick-colored-egg} task with instructions such as \textit{``pick up the red egg''}; this task includes OOD settings with \emph{unseen colors} and \emph{unseen positions}. The third is an \textbf{egg-carton placement} task, where the robot places an egg into a specified cell of a $4\times4$ carton grid according to language instructions such as \textit{``line 2, column 4''}; the OOD setting contains \emph{unseen row-column combinations}. For tasks with multiple OOD settings, we report the average OOD performance.

\begin{table}[h]
\centering
\caption{Summary of real-world OOD experiments with StarVLA-$\alpha$. We report performance under in-domain (ID) and out-of-distribution (OOD) settings across three real-world tasks. For tasks with multiple OOD settings, we report the average OOD performance.}
\label{tab:real_ood_summary}
\small
\begin{adjustbox}{max width=\linewidth}
\begin{tabular}{l cc cc cc cc}
\toprule
\multirow{2}{*}{\textbf{Metric}} 
& \multicolumn{2}{c}{\textbf{Pick colored egg}} 
& \multicolumn{2}{c}{\textbf{Egg carton placement}} 
& \multicolumn{2}{c}{\textbf{Waste-sorting}} 
& \multicolumn{2}{c}{\textbf{Average}} \\
\cmidrule(lr){2-3} \cmidrule(lr){4-5} \cmidrule(lr){6-7} \cmidrule(lr){8-9}
& \textbf{ID} & \textbf{OOD} 
& \textbf{ID} & \textbf{OOD} 
& \textbf{ID} & \textbf{OOD} 
& \textbf{ID} & \textbf{OOD} \\
\midrule
Success rate (\%) 
& 77.1 & 75.0 
& 91.3 & 68.8 
& 87.5 & 85.0 
& 85.3 & 76.3 \\
\bottomrule
\end{tabular}
\end{adjustbox}
\end{table}

\mypara{Results.}
The results are summarized in Table~\ref{tab:real_ood_summary}. Overall, StarVLA-$\alpha$ remains robust across all three real-world tasks and maintains strong performance under multiple forms of distribution shift. Notably, although StarVLA-$\alpha$ is intentionally simple and does not rely on elaborate data engineering, heavy data augmentation, or task-specific training tricks, its OOD performance remains largely comparable to its IID performance across all tasks. On average, the success rate only drops from 85.3\% in the IID setting to 76.3\% in the OOD setting, indicating that the robustness of StarVLA-$\alpha$ mainly comes from the learned policy itself rather than from dataset-specific engineering.

In the waste-sorting task, the OOD performance on novel objects remains very close to the in-domain result (85.0\% vs.\ 87.5\%), suggesting that the model learns category-level grounding rather than merely memorizing object appearance. In the pick-colored-egg task, StarVLA-$\alpha$ also generalizes well to both unseen colors and unseen positions, achieving success rates of 68.0\% and 81.9\%, respectively. This result indicates that the model can reliably bind language attributes to visual instances while preserving spatial robustness under distribution shift. In the egg-carton placement task, StarVLA-$\alpha$ achieves strong in-domain performance and remains effective on unseen coordinate combinations, although compositional spatial generalization is comparatively more challenging than the other OOD settings. Even in this more difficult case, the OOD result remains reasonably competitive relative to the IID performance, further demonstrating the stability of the framework in practical real-world settings.

\mypara{Discussion.}
These real-world results complement the benchmark evaluations in the main paper. Beyond achieving strong performance on standardized benchmarks, StarVLA-$\alpha$ also demonstrates stable instruction following and nontrivial OOD robustness in practical robotic manipulation tasks. More importantly, this robustness is obtained with a simple and unified framework, without requiring sophisticated data curation pipelines, additional augmentation strategies, or complex task-specific engineering. The relatively small gap from IID to OOD suggests that StarVLA-$\alpha$ learns a transferable visuomotor policy with genuine generalization ability, instead of overfitting to the exact training distribution. Taken together, these experiments further support our main conclusion that a simple VLM-based policy can already provide strong real-world generalization without additional architectural complexity.

\section{Full Benchmark Results}
\label{sec:full_results}

Due to space limitations in the main paper, we only report benchmark-level average results in the main text. In this appendix, we provide the full task-level results for several benchmarks to facilitate more detailed comparison and reproducibility. Specifically, we include detailed results for \textbf{SimplerEnv}, \textbf{RoboTwin-2.0}, and \textbf{RoboCasa-GR1}. All results follow the official evaluation protocols of each benchmark.

\subsection{SimplerEnv}

Table~\ref{tab:simplerenv_widowx_full} reports the detailed results on the SimplerEnv benchmark under the WidowX robot with the Visual Matching setting. The benchmark contains four manipulation tasks, and we report the success rate for each task together with the average performance. Results are shown for representative prior VLA methods as well as our StarVLA-$\alpha$ variants with different action heads and backbones.

\begin{table}[ht]
    \centering
    \caption{\textbf{Detailed results on SimplerEnv under the WidowX robot (VM).} We report per-task success rates and the average across tasks for prior methods and StarVLA-$\alpha$ variants.}
    \label{tab:simplerenv_widowx_full}
    \begin{adjustbox}{width=\linewidth}
    \begin{tabular}{l l c c c c c}
    \toprule
    \multicolumn{1}{c}{\textbf{WidowX Robot}} & \textbf{Method}
      & \makecell[c]{\textbf{Put Spoon} \\ \textbf{on Towel}} 
      & \makecell[c]{\textbf{Put Carrot} \\ \textbf{on Plate}} 
      & \makecell[c]{\textbf{Stack Green Block} \\ \textbf{on Yellow Block}} 
      & \makecell[c]{\textbf{Put Eggplant} \\ \textbf{in Yellow Basket}} 
      & \textbf{Average} \\
    \midrule
    \multirow{10}{*}{\makecell[l]{Visual\\Matching}}
      & RT-1-X~\cite{RT-1}  & 0.0  & 4.2  & 0.0  & 0.0  & 1.1 \\
      & Octo-Base~\cite{octo}    & 15.8 & 12.5 & 0.0  & 41.7 & 17.5 \\
      & Octo-Small~\cite{octo}  & 41.7 & 8.2  & 0.0  & 56.7 & 26.7 \\
      & OpenVLA~\cite{openvla}     & 4.2  & 0.0  & 0.0  & 12.5 & 4.2 \\
      & {CogACT}~\cite{cogact}              & {71.7}   & {50.8} & {15.0} & 67.5 & {51.3} \\
      & {SpatialVLA}~\cite{spatialvla}              & 16.7   & 25.0 & 29.2 & \textbf{100.0} & 42.7 \\

    & {$\pi_0$}~\cite{pi_0}              & {29.1} & {0.0}  & {16.6} & {62.5} & 27.1 \\
    & $\pi_0$-FAST~\cite{pertsch2025fast}  	& 29.1 & 21.9 & 10.8 & 66.6 & 48.3 \\
    & GR00T N1.5~\cite{bjorck2025gr00t}   & {75.3} & {54.3} & \textbf{{57.0}} & 61.3 & {61.9} \\ 
    & Magma~\cite{magma}  & 37.5 & 31.0 & 12.7  & 60.5 & 35.8 \\
    % \rowcolor{gray!10}
    \cmidrule(lr){2-7}
    & \textbf{StarVLA-$\boldsymbol{\alpha}$(Specialist)}  & \textbf{90.3} & 38.5 & 29.7 & \textbf{100} & 64.6 \\
    &\textbf{StarVLA-$\boldsymbol{\alpha}$(Generalist)}  & 79.7 & \textbf{59.8} & 22.8 & 98.5 & \textbf{65.2} \\
    \bottomrule
  \end{tabular}
  \end{adjustbox}
\end{table}

\begin{table}[ht!]
  \centering
\caption{Detailed results on the SimplerEnv Google Robot benchmark. Underlined scores indicate the best results excluding ours. Numbers are officially reported unless marked with $*$, which denotes our reimplementation.}
    \begin{adjustbox}{width=\linewidth}
  \begin{tabular}{ l l c c c c c } 
    \toprule
    \makecell[c]{Google\\Robot} 
    & \multicolumn{1}{c}{Models} 
    & \makecell[c]{Pick \\ Coke Can}
    & \makecell[c]{Move \\ Near} 
    & \makecell[c]{Open/Close \\ Drawer}
    & \makecell[c]{Open Top Drawer \\ and Place Apple} 
    & Avg \\
    \midrule
    \multirow{10}{*}{\makecell[l]{Visual\\Matching}}
      & RT-1~\cite{RT-1}      & 85.7 & 44.2 & \textbf{{73.0}} &  6.5 & 52.4 \\
      & RT-1-X~\cite{open_x_embodiment}  & 56.7 & 31.7 & 59.7 & 21.3 & 42.4 \\
      & RT-2-X~\cite{RT-2}  & 78.7 & 77.9 & 25.0 &  3.7 & 46.3 \\
      & OpenVLA~\cite{openvla}    & 18.0 & 56.3 & 63.0 &  0.0 & 34.3 \\
      & {CogACT}~\cite{cogact}      & {91.3} & {85.0} & {71.8} & {50.9} & 74.8 \\
       & SpatialVLA~\cite{spatialvla} 	& 86.0 &	77.9 & 	57.4 &	- & {75.1} \\
      & $\pi_0$~\cite{pi_0}	 &  72.7  & 65.3 & 	38.3 & 	- &  58.8  \\
 & $\pi_0$-FAST~\cite{pertsch2025fast}  & 	75.3 & 	67.5 & 	42.9  & -  & 	61.9 \\
        & GR00T N1.5$^*$~\cite{bjorck2025gr00t}  & 51.7 & 54.0 & 27.8 & 7.4 & 35.2 \\ 
        & Magma~\cite{magma}   & 83.7& 65.4  &  56.0 &  6.4 & 52.9 \\
        \cmidrule(lr){2-7}

    &   \textbf{StarVLA-$\boldsymbol{\alpha}$ (Specialist)}   &  \textbf{95.3} & {75.0} & {68.8} & {66.1} & {76.0} \\
    
    & \textbf{StarVLA-$\boldsymbol{\alpha}$ (Generalist)}      & {90.1} & \textbf{82.6} & {56.3} & \textbf{68.7} & \textbf{74.3} \\
    \midrule
    \multirow{10}{*}{\makecell[l]{Variant\\Aggregation}}
      & RT-1~\cite{RT-1}      & {89.8} & 50.0 & 32.3 &  2.6 & 43.7 \\
      & RT-1-X~\cite{open_x_embodiment}  & 49.0 & 32.3 & 29.4 & 10.1 & 30.2 \\
      & RT-2-X~\cite{RT-2}  & 82.3 & 79.2 & 35.3 & 20.6 & 54.4 \\
      % & Octo-Base~\cite{octo} &  0.6 &  3.1 &  1.1 &  0.0 &  1.2 \\
      & OpenVLA~\cite{openvla}    & 60.8 & 67.7 & 28.8 &  0.0 & 39.3 \\
      & {CogACT}~\cite{cogact}      & {89.6} & \textbf{{80.8}} & 28.3 & {46.6} & {61.3} \\
       & SpatialVLA~\cite{spatialvla} 	& 88.0 & {82.5} & 	{41.8} &	- & {70.7} \\
        & $\pi_0$~\cite{pi_0}	 & 75.2 &	63.7 &	25.6 &	- & 54.8 \\
        & $\pi_0$-FAST~\cite{pertsch2025fast} 	& 77.6 &	68.2 &	31.3 & - &	59.0 \\
        & GR00T N1.5~\cite{bjorck2025gr00t}  & 69.3 & 68.7 & 35.8 & 4.0 & 44.5 \\
         & Magma~\cite{magma}  & 68.8 & 65.7  & {53.4} & 18.5 & 51.6 \\
        \cmidrule(lr){2-7}
        & \textbf{StarVLA-$\boldsymbol{\alpha}$ (Specialist)} & \textbf{91.3} & {75.1} & {55.0} & \textbf{59.4} & \textbf{70.2} \\
    & \textbf{StarVLA-$\boldsymbol{\alpha}$ (Generalist)} & {88.8} & \textbf{{78.8}} & \textbf{56.1} & {55.5} & {69.8} 
    \\
      \bottomrule
  \end{tabular}
  \end{adjustbox}
  \label{tab:simpler-google} 
\end{table}

\subsection{RoboTwin-2.0}

Table~\ref{tab:robotwin-cleanvla-full} presents the full task-level results of StarVLA-$\alpha$ on RoboTwin-2.0. The benchmark consists of 50 dual-arm manipulation tasks, each evaluated under both Easy and Hard settings. We report the success rate for each task as well as the overall average across all tasks.

\begin{table*}[h]
\centering
\caption{\textbf{Detailed results of StarVLA-$\alpha$ on RoboTwin 2.0 under specialist setting.} We report success rates for each task under the Easy and Hard settings.}
\label{tab:robotwin-cleanvla-full}
\begin{adjustbox}{max width=\linewidth}
\small
\begin{tabular}{lcc|lcc|lcc}
\toprule
\multicolumn{9}{c}{RoboTwin-2.0} \\
\midrule
Task & Easy & Hard & Task & Easy & Hard & Task & Easy & Hard \\
\midrule

Adjust Bottle & 100 & 99 &
Open Microwave & 28 & 39 &
Place Object Stand & 99 & 98 \\

Beat Block Hammer & 93 & 92 &
Pick Diverse Bottles & 87 & 86 &
Place Phone Stand & 86 & 95 \\

Blocks Ranking RGB & 99 & 98 &
Pick Dual Bottles & 91 & 93 &
Place Shoe & 96 & 100 \\

Blocks Ranking Size & 79 & 80 &
Place A2B Left & 90 & 95 &
Press Stapler & 99 & 96 \\

Click Alarmclock & 58 & 51 &
Place A2B Right & 88 & 95 &
Put Bottles Dustbin & 90 & 85 \\

Click Bell & 23 & 27 &
Place Bread Basket & 91 & 78 &
Put Object Cabinet & 89 & 91 \\

Dump Bin Bigbin & 91 & 94 &
Place Bread Skillet & 89 & 80 &
Rotate QRcode & 88 & 90 \\

Grab Roller & 100 & 100 &
Place Burger Fries & 100 & 100 &
Scan Object & 94 & 91 \\

Handover Block & 97 & 93 &
Place Can Basket & 75 & 75 &
Shake Horizontally & 100 & 100 \\

Handover Mic & 98 & 96 &
Place Cans Plasticbox & 100 & 99 &
Shake Bottle & 100 & 100 \\

Hanging Mug & 34 & 29 &
Place Container Plate & 99 & 99 &
Stack Blocks Three & 94 & 86 \\

Lift Pot & 100 & 100 &
Place Dual Shoes & 91 & 89 &
Stack Blocks Two & 100 & 100 \\

Move Can Pot & 91 & 90 &
Place Empty Cup & 100 & 100 &
Stack Bowls Three & 95 & 91 \\

Move Pillbottle Pad & 98 & 100 &
Place Fan & 94 & 95 &
Stack Bowls Two & 99 & 100 \\

Move Playingcard Away & 100 & 98 &
Place Mouse Pad & 87 & 94 &
Stamp Seal & 86 & 90 \\

Move Stapler Pad & 74 & 90 &
Place Object Basket & 93 & 94 &
Turn Switch & 65 & 62 \\

Open Laptop & 98 & 100 &
Place Object Scale & 93 & 93 &
\textbf{Average} & \textbf{88.2 }& \textbf{88.3} \\

\bottomrule
\end{tabular}
\end{adjustbox}
\end{table*}

\subsection{RoboCasa-GR1}

Table~\ref{tab:robocasa-cleanvla-updated} shows the detailed evaluation results on the RoboCasa-GR1 tabletop benchmark. The benchmark contains 24 tasks, and each model is trained jointly across all tasks. We report the success rate of each task together with the overall average performance.

\begin{table}[htbp]
\centering
\caption{\textbf{Evaluation results on the RoboCasa-GR1 tabletop benchmark.} A single model is trained jointly on all 24 tasks, and results are reported over 200 rollouts per task.}
\label{tab:robocasa-cleanvla-updated}
\begin{adjustbox}{max width=\textwidth}
\begin{tabular}{lc | c | c}
\toprule
\textbf{Task} & \textbf{GR00T-N1.6} & \textbf{StarVLA-$\boldsymbol{\alpha}$(Specialist)} & \textbf{StarVLA-$\boldsymbol{\alpha}$(Generalist)} \\
\midrule
PnPBottleToCabinetClose & 51.5 & 35.0 & \textbf{52.0} \\
PnPCanToDrawerClose & 13.0 & 81.0 & \textbf{86.0} \\
PnPCupToDrawerClose & 8.5 & \textbf{50.0} & 38.0 \\
PnPMilkToMicrowaveClose & 14.0 & 49.0 & \textbf{56.0} \\
PnPPotatoToMicrowaveClose & 41.5 & 37.0 & \textbf{46.0} \\
PnPWineToCabinetClose & 16.5 & 42.0 & \textbf{46.0} \\
\addlinespace
PnPNovelFromCuttingboardToBasket & \textbf{58.0} & 55.0 & 56.0 \\
PnPNovelFromCuttingboardToCardboardbox & 46.5 & 45.0 & \textbf{48.0} \\
PnPNovelFromCuttingboardToPan & 68.5 & 75.0 & \textbf{80.0} \\
PnPNovelFromCuttingboardToPot & \textbf{65.0} & 59.0 & 60.0 \\
PnPNovelFromCuttingboardToTieredbasket & \textbf{46.5} & 43.0 & 42.0 \\
\addlinespace
PnPNovelFromPlacematToBasket & 58.5 & 38.0 & \textbf{60.0} \\
PnPNovelFromPlacematToBowl & 57.5 & 63.0 & \textbf{74.0} \\
PnPNovelFromPlacematToPlate & 63.0 & 57.0 & \textbf{74.0} \\
PnPNovelFromPlacematToTieredshelf & 28.5 & \textbf{29.0} & 28.0 \\
\addlinespace
PnPNovelFromPlateToBowl & 57.0 & 65.0 & \textbf{72.0} \\
PnPNovelFromPlateToCardboardbox & 43.5 & \textbf{55.0} & 44.0 \\
PnPNovelFromPlateToPan & 51.0 & \textbf{71.0} & 70.0 \\
PnPNovelFromPlateToPlate & \textbf{78.7} & 73.0 & 74.0 \\
\addlinespace
PnPNovelFromTrayToCardboardbox & \textbf{51.5} & 49.0 & 48.0 \\
PnPNovelFromTrayToPlate & 71.0 & 61.0 & \textbf{72.0} \\
PnPNovelFromTrayToPot & 64.5 & \textbf{67.0} & \textbf{67.0} \\
PnPNovelFromTrayToTieredbasket & 57.0 & \textbf{59.0} & 58.0 \\
PnPNovelFromTrayToTieredshelf & 31.5 & \textbf{33.0} & 24.0 \\
\midrule
\textbf{Average} & {47.6} & {53.8} & \textbf{57.3} \\
\bottomrule
\end{tabular}
\end{adjustbox}
\end{table}

\clearpage
\clearpage

\section{Result Visualization Across Simulation Benchmarks}
\label{sec:vis_all_benchmarks}

To provide a clearer overview of the evaluation environments used in this work, we visualize representative scenes from all benchmarks considered in the paper. These visualizations help illustrate the diversity of robot embodiments, task layouts, and manipulation scenarios across the different benchmarks. Since our experiments span multiple datasets with different robot platforms and environments, these figures provide an intuitive understanding of the visual observations encountered by the policies during evaluation.

\begin{figure}[ht]
    \centering
    \includegraphics[width=1\linewidth]{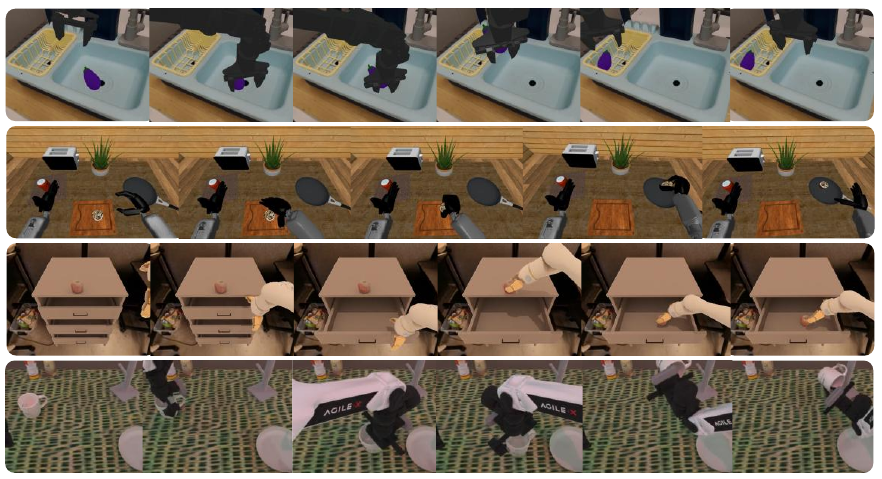}
    \caption{\textbf{Result visualization across simulation benchmarks}. From top to bottom: SimplerEnv WidowX, RoboCasa-GR1, SimplerEnv Google Robot, and RoboTwin 2.0 (Hard).}
    \label{fig:vis-simbench}
\end{figure}

Figure~\ref{fig:vis-simbench} presents example frames from the simulation benchmarks used in our experiments. From top to bottom, the figure shows scenes from SimplerEnv with the WidowX robot, RoboCasa-GR1, SimplerEnv with the Google Robot embodiment, and RoboTwin 2.0 under the Hard setting. These environments cover a wide range of manipulation settings, including single-arm tabletop manipulation, humanoid-style interaction scenarios, and dual-arm coordination tasks. Despite the differences in embodiment and environment structure, all benchmarks follow language-conditioned manipulation protocols and share similar RGB-based observations.

\begin{figure}[ht]
    \centering
    \includegraphics[width=1\linewidth]{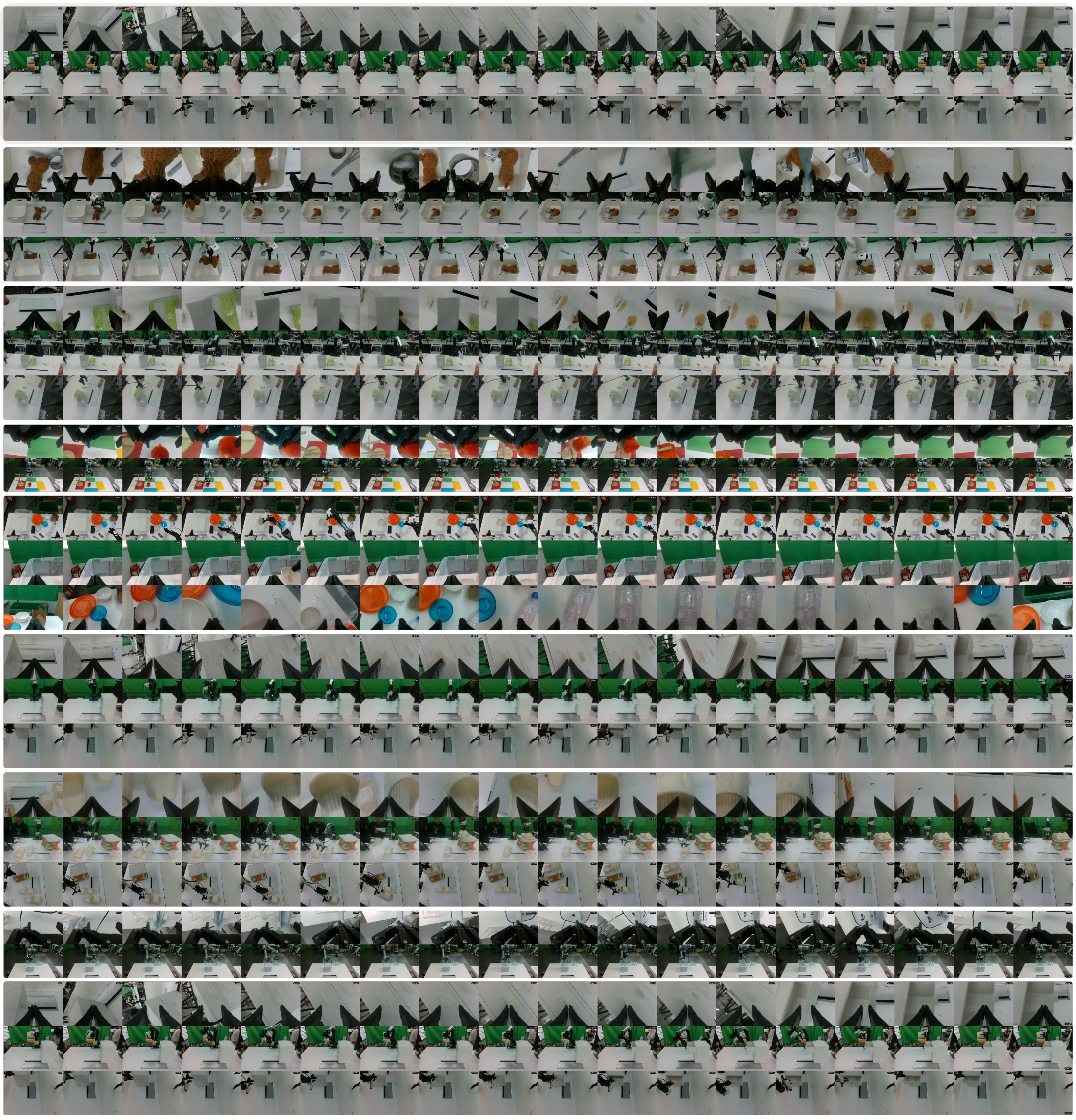}
    \caption{\textbf{Result visualization of large-scale real-world benchmark on RoboChallenge}. See supplementary webpage for more videos.}    \label{fig:vis-robochallenge}
\end{figure}

Figure~\ref{fig:vis-robochallenge} shows representative scenes from the RoboChallenge real-world benchmark used in our evaluation. Compared with simulation environments, RoboChallenge introduces additional challenges such as real-world sensing noise, lighting variations, and execution uncertainty. These visualizations provide an example of the physical robot setup and task environment used for real-world evaluation.

\begin{figure}[ht]
    \centering
    \includegraphics[width=1\linewidth]{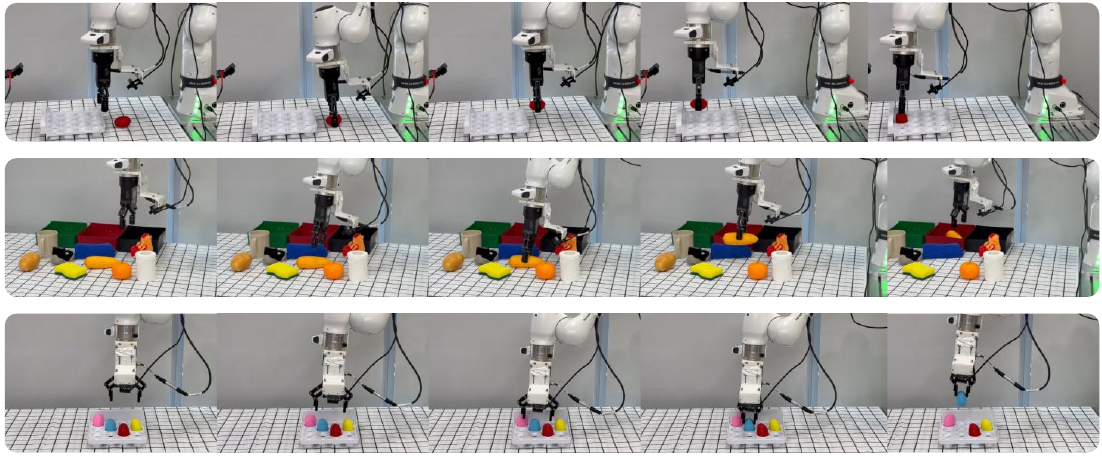}
    \caption{\textbf{Real-world deployment tasks on Franka Research 3.} From top to bottom: egg-carton placement, waste sorting and colored egg picking.}
    \label{fig:vis-real-world}
\end{figure}

Figure~\ref{fig:vis-real-world} further presents representative scenes from our real-world deployment experiments on the Franka Research 3 robot. Different from RoboChallenge, which emphasizes standardized large-scale evaluation across hosted robot platforms, these experiments are designed to study real-world instruction following and out-of-distribution generalization under our own deployment setup. The figure illustrates the physical tabletop environment, camera viewpoints, and representative manipulation tasks used in our evaluation, including waste sorting, colored egg picking, and egg-carton placement. Compared with benchmark-based evaluation, these real-world tasks involve more unconstrained object appearances, spatial variations, and language-conditioned target specifications, providing a complementary view of how StarVLA-$\alpha$ behaves in practical deployment scenarios.

\section{Robustness Evaluation on LIBERO-Plus} \label{sec:libero_plus}
% 介绍 LIBERO-Plus
LIBERO-Plus is an extended benchmark built upon the standard LIBERO dataset to evaluate the robustness of robot manipulation models under diverse perturbations. 
It introduces variations in camera viewpoint, robot configuration, language instructions, lighting conditions, background clutter, sensor noise, and object layout.

In our evaluation, we strictly follow the official setup: all models are trained on the standard LIBERO training data and directly evaluated on LIBERO-Plus without any additional adaptation. Therefore, this benchmark measures whether the policy learned from standard LIBERO can transfer to perturbed environments, rather than whether it can fit a specific robustness-oriented training distribution.

\begin{table}[ht]
\centering
\caption{\textbf{Performance comparison on the LIBERO-Plus benchmark under perturbations in camera, robot, language, lighting, background, noise, and layout.} All models are trained on standard LIBERO and evaluated on LIBERO-Plus, and the best score in each column is shown in bold.}
\label{tab:libero-plus-performance}
\small
\begin{adjustbox}{max width=\textwidth}
\begin{tabular}{lcccccccc}
\toprule
\textbf{Model} & \textbf{Camera} & \textbf{Robot} & \textbf{Language} & \textbf{Light} & \textbf{Background} & \textbf{Noise} & \textbf{Layout} & \textbf{Total} \\
\midrule
OpenVLA~\cite{openvla}         & 0.8  & 3.5  & 23.0 & 8.1  & 34.8 & 15.2 & 28.5 & 15.6 \\
OpenVLA-OFT~\cite{openvla-oft}     & 56.4 & 31.9 & 79.5 & 88.7 & 93.3 & 75.8 & 74.2 & 69.6 \\
NORA~\cite{hung2025norasmallopensourcedgeneralist}            & 2.2  & 37.0 & 65.1 & 45.7 & 58.6 & 12.8 & 62.1 & 39.0 \\
WorldVLA~\cite{worldvla}        & 0.1  & 27.9 & 41.6 & 43.7 & 17.1 & 10.9 & 38.0 & 25.0 \\
UniVLA~\cite{bu2025univla}          & 1.8  & 46.2 & 69.6 & 69.0 & 81.0 & 21.2 & 31.9 & 43.9 \\
$\pi_0$~\cite{pi_0}         & 13.8 & 6.0  & 58.8 & 85.0 & 81.4 & 79.0 & 68.9 & 53.6 \\
$\pi_0$-Fast~\cite{pertsch2025fast}    & \textbf{65.1} & 21.6 & 61.0 & 73.2 & 73.2 & 74.4 & 68.8 & 61.6 \\
RIPT-VLA~\cite{tan2025interactiveposttrainingvisionlanguageactionmodels}        & 55.2 & 31.2 & 77.6 & 88.4 & 91.6 & 73.5 & 74.2 & 68.4 \\
\midrule
StarVLA-$\alpha$ (Specialist) & 48.7 & {63.4} & \textbf{86.8} & {95.8} & {94.6} & 75.0 & \textbf{80.2} & 77.8 \\ 
StarVLA-$\alpha$ (Generalist) & {52.5} & \textbf{64.3} & 86.2 & \textbf{97.8} & \textbf{98.1} & \textbf{80.2} & 79.1 & \textbf{79.7} \\ 
\bottomrule
\end{tabular}
\end{adjustbox}
\label{tab:performance}
\end{table}

Table~\ref{tab:libero-plus-performance} shows that StarVLA-$\alpha$ transfers well from standard LIBERO to LIBERO-Plus under diverse perturbations. Both specialist and generalist variants remain robust and outperform prior baselines, despite being trained only on standard LIBERO without any benchmark-specific robustness augmentation.

Notably, StarVLA-$\alpha$ performs consistently well under language, lighting, background, and layout perturbations, suggesting that a strong VLM backbone provides robust multimodal representations. The generalist model is also competitive with, and sometimes slightly better than, the specialist model, supporting our main finding that joint training across benchmarks can improve robustness. Overall, these results show that strong initialization and a unified training pipeline already yield substantial robustness without extra architectural complexity.

\end{document}